\documentclass[letterpaper, 10 pt, conference]{ieeeconf}  

\IEEEoverridecommandlockouts                              

\usepackage{tikz}
\usepackage{adjustbox}
\usepackage{algorithm}               
\usepackage{algorithmic}             
\usepackage{booktabs}
\usepackage{epstopdf}
\usepackage{enumerate}
\usepackage{pdfpages}
\usepackage{afterpage}
\usepackage{color, colortbl}
\usepackage{footmisc}
\usepackage{etoolbox}
\usepackage{tablefootnote}
\usepackage[flushleft]{threeparttable}
\usepackage{graphicx}
\usepackage{subfig}
\usepackage{caption}
\usepackage{multirow}
\usepackage{verbatim}
\usepackage{footnote}
\usepackage{url}
\usepackage{amsmath,amssymb} 
\usepackage{pifont}
\newcommand{\cmark}{\ding{51}}%
\newcommand{\xmark}{\ding{55}}%

\title{\LARGE \bf
IRS: A Large Naturalistic Indoor Robotics Stereo Dataset to Train Deep Models for Disparity and Surface Normal Estimation
}

\author{Qiang Wang$^{1,*}$, Shizhen Zheng$^{1,*}$, Qingsong Yan$^{2,*}$, Fei Deng$^{2}$, Kaiyong Zhao$^{1,\dagger}$, Xiaowen Chu$^{1,\dagger}$ 
\thanks{$^{*}$Authors have contributed equally.}
\thanks{$^{\dagger}$Corresponding authors.}
\thanks{$^{1}$Department of Computer Science, Hong Kong Baptist University, {\tt\small \{qiangwang,szzheng,kyzhao,chxw@comp.hkbu.edu.hk\}}}
\thanks{$^{2}$School of Geodesy and Geomatics, Wuhan University, {\tt\small \{yanqs\_whu@whu.edu.cn, fdeng@sgg.whu.edu.cn\}}}
}

\begin{document}

\maketitle
\thispagestyle{empty}
\pagestyle{empty}

\begin{abstract}
Indoor robotics localization, navigation and interaction heavily rely on scene understanding and reconstruction. Compared to monocular vision which usually does not explicitly introduce any geometrical constraint, stereo vision based schemes are more promising and robust to produce accurate geometrical information, such as surface normal and depth/disparity. Besides, deep learning models trained with large-scale datasets have shown their superior performance in many stereo vision tasks. However, existing stereo datasets rarely contain the high-quality surface normal and disparity ground truth, which hardly satisfy the demand of training a prospective deep model for indoor scenes.

To this end, we introduce a large-scale synthetic but naturalistic indoor robotics stereo (IRS) dataset with over 100K stereo RGB images and high-quality surface normal and disparity maps. Leveraging the advanced rendering techniques of our customized rendering engine, the dataset is considerably close to the real-world captured images and covers several visual effects, such as brightness changes, light reflection/transmission, lens flare, vivid shadow, etc. We compare the data distribution of IRS with existing stereo datasets to illustrate the typical visual attributes of indoor scenes. Besides, we present DTN-Net, a two-stage deep model for surface normal estimation. Extensive experiments show the advantages and effectiveness of IRS in training deep models for disparity estimation, and DTN-Net provides state-of-the-art results for normal estimation compared to existing methods.
\end{abstract}

\section{Introduction}
Indoor scene understanding and reconstruction are central to many robotics applications, such as robot localization, navigation, and interaction. Despite the attractive cost and availability, monocular vision does not explicitly introduce any geometrical constraint. On the contrary, stereo vision leverages the advantage of cross-reference between the left and the right view, and usually shows greater performance and robustness in geometric information inference, such as surface normal and disparity/depth estimation. Recent advances \cite{fast2018,flownet3,ganet,psmnet,groupnet,geonet_normal,ednet2021} in those tasks have shown that deep neural network (DNN) can significantly improve the estimation quality. However, the success of DNN requires large scale and high-quality labelled datasets, which are still lacking in stereo vision studies.

Surface normal and disparity/depth are two core information in 3D geometry understanding since they can determine the position and orientation of an object in the space. In addition, they also have strong knowledge relation. On the one hand, surface normal is determined by local surface tangent plane of neighboring 3D points, which can be calculated from depth; on the other hand, the orientation of the plane constructed by the depth is constrained by the surface normal. This knowledge has been used in \cite{bleyer2011patchmatch,zhang2017robust,scharstein2017semi,normal2020cvpr,geonet2020parmi} to jointly optimize the prediction quality. 

Existing studies \cite{scharstein2003high,scharstein2007learning,scharstein2014high,kitti2012} proposed stereo datasets collected by real sensing hardware, which contributed a lot to stereo vision research. However, they typically have only a small number of samples and lack complete and dense ground-truth of surface normal and disparity. Recent work in \cite{butler2012naturalistic,mayer2016large,Zhang_2017_CVPR,hu2021orstereo,tartanair2020iros} leveraged synthetic technology to generate sufficiently large data volume for DNN training. However, there are two main drawbacks of them. First, few of stereo vision datasets contains both high-quality disparity and surface normal ground truth. Second, due to the limitation of the rendering systems, their RGB images are usually noisy and not realistic enough in terms of brightness variation, light reflection/transmission, indirect shadows, bloom, lens flare, etc. It has been shown that the existing state-of-the-art deep models \cite{psmnet,mayer2016large,aanet2019} trained on these synthetic datasets did not work well on the complicated real-world indoor scenes.




To this end, we propose IRS, a large scale synthetic stereo dataset for indoor robotics applications, which is generated by a customized advanced rendering engine. We conduct quantitative analysis and deep model training experiments to verify the effectiveness of learning indoor geometrical information from IRS. We also make our dataset as well as the model training scripts publicly available\footnote{\url{https://github.com/HKBU-HPML/IRS}}. Our contributions are summarized as follows:
\begin{itemize}
	\item We present a large synthetic indoor robotics stereo dataset, namely IRS, generated by a customized version of Unreal Engine (UE4) with originally implemented plug-ins. Our dataset contains over 100K of stereo images as well as their complete surface normal and disparity ground truth. With the advanced rendering techniques of UE4, the vision attributes of the real-world physical environment can be well simulated.
	\item We conduct quantitative analysis to compare IRS with some existing stereo datasets. We show that IRS covers common visual attributes of indoor scenes, including lightness variation and camera vision range changes. 
	\item Compared to the existing model trained on FlyingThings3D, another stereo dataset, the same network architecture trained with IRS yields higher accuracy and better generalization on existing stereo datasets. Besides, we present DTN-Net to predict the surface normal maps for indoor scene stereo images. DTN-Net predicts better surface normal for indoor scenes than the existing RGB-based and depth-fusion models. 
\end{itemize}

The rest of the paper is organized as follows. We introduce the existing studies on stereo datasets in Section \ref{sec:related_work}. Then we present the methodology and implementation of generating IRS dataset in Section \ref{sec:data_generation}. Section \ref{sec:quan_analysis} presents a quantitative comparison between IRS and existing stereo datasets. In Section \ref{sec:exp}, we present our performance evaluation to illustrate the effectiveness of IRS on  stereo data. We finally conclude the paper in Section \ref{sec:conclusion}.
\section{Related Work}\label{sec:related_work}
Indoor scene understanding is vital to many robotics applications. Monocular vision mainly focuses on core scene understanding tasks such as object detection \cite{Girshick_2015_ICCV,ssd} and semantics segmentation \cite{Zhang_2018_CVPR,Long_2015_CVPR}. Recent popular monocular datasets include SUN3D \cite{sun3d2013}, ScanNet \cite{scannet2017}, Matterport3D \cite{matterport2017}, SceneNet RGB-D \cite{scenenet2017} and InteriorNet \cite{interiornet2018}. Stereo vision is popular to infer the spatial geometric information, including surface normal and depth/disparity, which play a significant role in the fields of autonomous driving and indoor robots. Compared to monocular disparity estimation and normal estimation which rely on the prior information of the scene and lack geometric constraints, stereo vision can combine the prior information and geometry information together to give better estimated results  \cite{bleyer2011patchmatch,zhang2017robust,scharstein2017semi}. 

Before the popularity of DNN models on solving big data training tasks, the main usage of the datasets is to evaluate different algorithms, which means that the data complexity and variety is more important than the data volume. MiddleBury is a well-known and frequently updated real world stereo dataset  \cite{scharstein2002taxonomy,scharstein2003high,scharstein2007learning,hirschmuller2007evaluation,scharstein2014high}, which takes the resolution, brightness, exposure and many other uncommon factors into consideration to improve the complexities of the dataset and provides dense per-pixel disparities of the scenes. 

With the rapid development of deep learning methods \cite{survey2020stereo}, traditional datasets can no longer meet the demand of training a deep model of decent robustness and generalization. 
Moreover, recent automated machine learning (autoML) \cite{xin2021automl,leastereo2020,cai2020once} techniques also require numerous training data to boost the model quality.
Datasets with larger volume and more complicated distributions have received more attention. 

Sintel \cite{butler2012naturalistic} is a synthetic dataset based on open source 3D movies. It uses Blender 3D engine to render the scene and obtain corresponding depth information and fully considers the influence of various factors, like motion blur and defocus blur. It provides 1,064 stereo images with high-quality disparity maps and has been used to train effective networks on realistic data in \cite{mayer2016large,flownet3,fast2018}.

KITTI is a natural dataset for autonomous driving, which provides 394 pairs to train and 394 pairs to test. KITTI2012 \cite{kitti2012} released the real world captured stereo images on roads as well as their disparity maps of high sparsity obtained by Velodyne HDL-64E. KITTI2015 \cite{kitti2015} extends the dataset by modeling cars, which may exist some fitting errors.
\begin{table*}[!t]
	\centering
	\begin{threeparttable}
		\caption{The Comparison of Recent Stereo Datasets and Our IRS}
		\label{tab:analytical_compare}
		\begin{tabular}
			{|l|c|c|c|c|c|c|c|c|c|} \hline
			\textbf{Dataset} & MiddleBury\cite{scharstein2014high} & KITTI\cite{kitti2012}\cite{kitti2015} & DrivingStereo\cite{drivingstereo2019} & Sintel\cite{butler2012naturalistic} & Apollo \cite{huang2018apolloscape} & Scene Flow\cite{mayer2016large} & IRS(ours) \\ \hline
			
			\textbf{Synthetic(S) / Natural(N)} & N & N & N & S & N & S & S \\ \hline
			\textbf{Scene} & Lab & Road & Road & Outdoor & Road & Outdoor & Indoor \\ \hline
			\textbf{Resolution} & 2960x1942 & 1226x370 & 1242x375 & 1024x436 & 3130x960 & 960x540 & 960x540 \\ \hline
			\textbf{Training/Testing Data Size} & 23/10 & 194/194 & 200/200 & 1064/0 & 22390/0 & 35454/4370 & 84946/15079 \\ \hline
			\textbf{Density of Ground Truth} & 100\% & $\le$ 30\% & $\le$ 30\% & 100\% & 100\% & 100\% & 100\% \\ \hline
			\textbf{Number of Texture Types} & 1 & 1 & 1 & 1 & 1 & Multiple & Multiple \\ \hline
			\textbf{Surface Normal} & \xmark & \xmark & \xmark & \xmark & \xmark & \xmark & \cmark \\ \hline
			\textbf{Textureless Region} & $\overline{\times}$ & $\overline{\times}$ & $\overline{\times}$ & $\overline{\times}$ & $\overline{\times}$ & $\overline{\times}$ & $\overline{\surd}$ \\ \hline
			\textbf{Reflection \& Light} & $\overline{\surd}$ & $\overline{\times}$ & $\overline{\times}$ & $\overline{\times}$ & $\overline{\surd}$ & $\overline{\times}$ & $\overline{\surd}$ \\ \hline
		\end{tabular}
	\end{threeparttable}
\end{table*}

Scene Flow \cite{mayer2016large} has over 39,000 pairs to train the network, which is based on the 3D model provided by ShapeNet \cite{savva2015semantically} and the texture from Flicker. Unlike other synthetic datasets that based on the time-consuming 3D engine, the data of Scene Flow is constructed by random selection of scene backgrounds, 3d objects, and the textures on those objects.

Apollo Stereo \cite{huang2018apolloscape}, like KITTI, is also a dataset for autonomous driving, but it uses two VUX-1HAs to get more denser depth information, providing a total of 22,390 pairs for training. Its ground truth is acquired by accumulating 3D point clouds from Lidar and fitting 3D CAD models to individually moving cars. Another novel in-the-wild stereo image dataset is Holopix50k \cite{hua2020holopix50k}, which comprises 49,368 image pairs contributed by users of the Holopix$^{\text{TM}}$ mobile social platform. Holopix50k covers more outdoor environmental settings, including fire, snow, sunset and so on. This large variety of diverse scenarios can significantly improves the generalization of deep models. However, Holopix50k lacks both disparity and normal information, and thus cannot be used for supervised model training. 

Currently, stereo datasets are either object-centric, such as Sintel and Scene Flow, where objects usually appear in the center of the image; or for autonomous driving, of which the main scenes are roads. With the development of indoor robots, stereo vision solutions for indoor scenes is becoming increasingly important. This paper presents a large scale indoor robotics stereo dataset, called IRS, generated by our customized Unreal Engine plug-in. The samples in IRS are close to the real world captured pictures in the aspects of material texture, light reflection and transmission, shadows, bloom and len flare. It also provides high-quality and dense labelled ground truth of surface normal and disparity. Table \ref{tab:analytical_compare} compares our IRS dataset and those existing stereo datasets. IRS provides over 100,000 samples with various texture types and per-pixel depth information. In addition, IRS provides dense surface normal labels and covers indoor scenarios of some special attributes, such as textureless region and light reflection, while the others only support some or even none of them.

\section{Data Generation}\label{sec:data_generation}
\begin{figure}[htbp]
	\centering
	\includegraphics[width=0.96\linewidth]{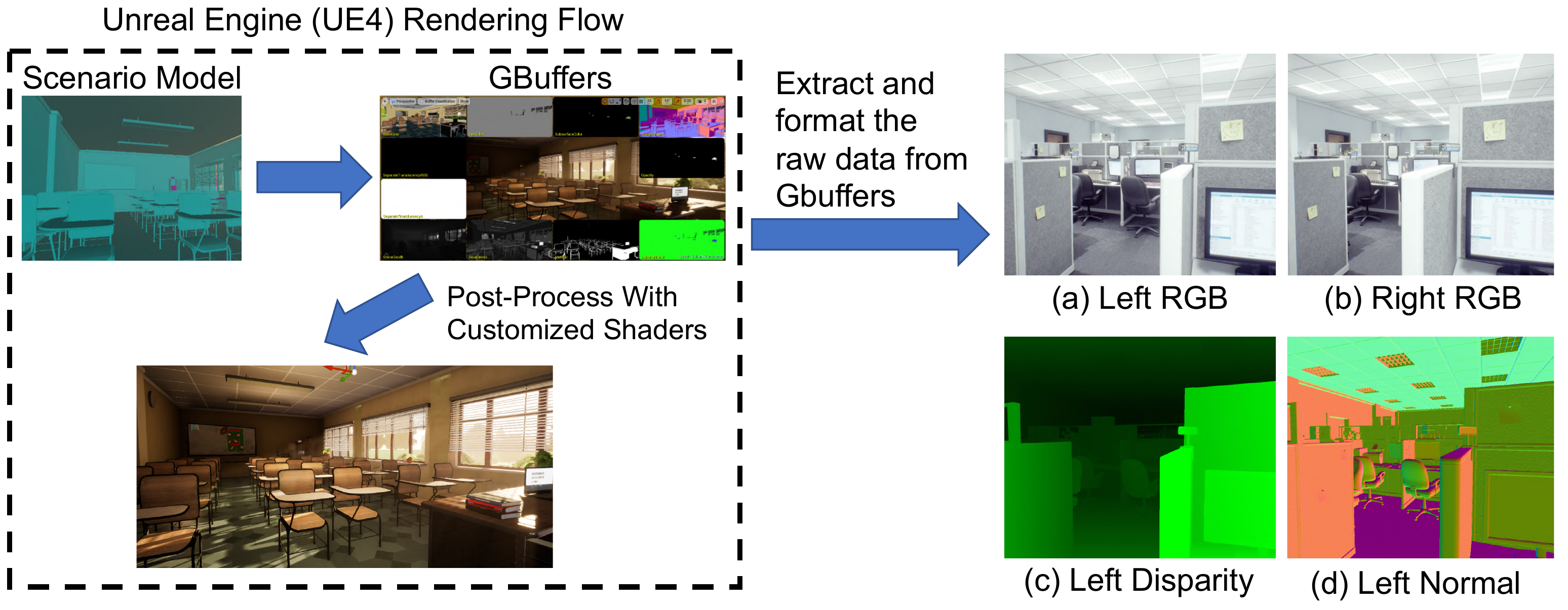}
	\caption{Data sample generated by our customized UE4 plug-in. The left part shows the original rendering flow, and the right part shows how the plug-in works to produce the IRS data samples.}
	\label{fig:ue4_render}
\end{figure}

The Unreal Engine 4 (UE4) is a powerful 3D creation software for photo-real visuals and immersive experiences. The left part of Fig. \ref{fig:ue4_render} demonstrates how UE4 renders the target scenario. It firstly establishes a 3D model which describes the 3D shape information of the objects in the scene. Then the deferred shading uses the concept of a GBuffer (Geometry Buffer) which stores the information about the geometry such as the world normal, base color, roughness, etc. Finally, UE4 samples these buffers when lighting is calculated to determine the final shading. By default, the rendering flow is sealed and cannot be split. 
Thus, we modified the pipeline of UE4's internal rendering engine and customized a plug-in. The plug-in extracts the raw data from the GBuffers and transforms them into the desired formats, which are a stereo pair of RGB images, the disparity map and the surface normal map in our study, as shown in the right part of Fig. \ref{fig:ue4_render}. The high-quality dense ground truth of surface normal and disparity maps are critical to train a prospective machine learning model.

\begin{figure}[htbp]
	\centering
	\subfloat[Home]
	{
		\includegraphics[width=0.3\linewidth]{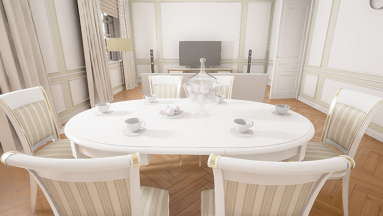}\label{fig:home}
	}
	\subfloat[Office]
	{
		\includegraphics[width=0.3\linewidth]{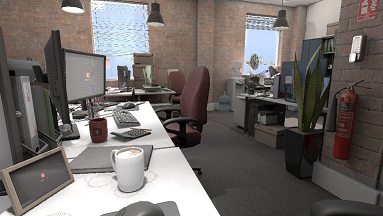}\label{fig:office}
	}
	\subfloat[Restaurant]
	{
		\includegraphics[width=0.3\linewidth]{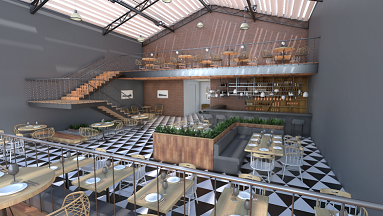}\label{fig:restaurant}
	}
	\qquad
	\subfloat[Dark]
	{
		\includegraphics[width=0.3\linewidth]{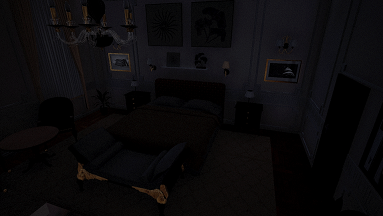}\label{fig:darkness}
	}
	\subfloat[Normal Light]
	{
		\includegraphics[width=0.3\linewidth]{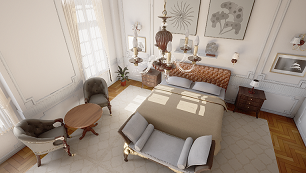}\label{fig:normal_light}
	}
	\subfloat[Over Exposure]
	{
		\includegraphics[width=0.3\linewidth]{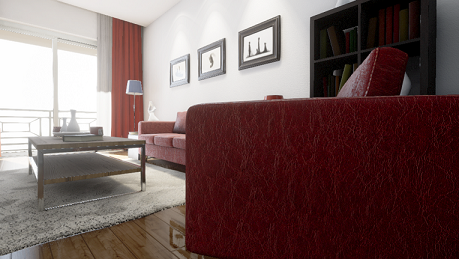}\label{fig:over_exposure}
	}
	\qquad
	\subfloat[Glass]
	{
		\includegraphics[width=0.3\linewidth]{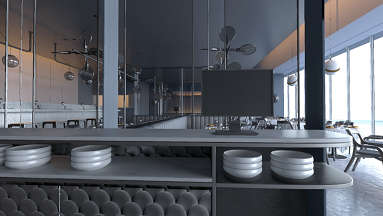}\label{fig:glass}
	}
	\subfloat[Mirror]
	{
		\includegraphics[width=0.3\linewidth]{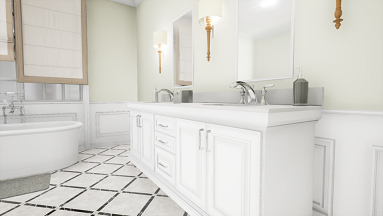}\label{fig:mirror}
	}
	\subfloat[Metal]
	{
		\includegraphics[width=0.3\linewidth]{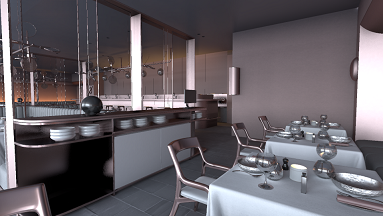}\label{fig:metal}
	}
	\caption{The first row shows different indoor scenes. The second row shows different lightness environments. The third row shows the light transmission/reflection effects of different materials.}
	\label{fig:data_different}
\end{figure}

In our plug-in, there are two classes of configurable factors that help generate samples of various visual effects. The first class is the scene factors, such as layout, object types, surface materials, lighting, etc. The second class is the camera parameters, including the intrinsic ones and the extrinsic ones. To construct the IRS dataset, we chose totally 21 indoor scenes and adjusted different configurations for them to generate over 100,000 samples. The samples of IRS cover common factors in real-world indoor scenes, including highlights, light colors, over-exposure, shadows, dark environments, specular reflections, metal surfaces, noise, etc., as illustrated in Fig. \ref{fig:data_different}.
With the advanced rendering techniques, such as ambient occlusion, diffuse inter-reflection, etc., UE4 provides high simulation effects close to the real world captured images in terms of light and shadow effects, brightness variation, bloom and len flare, and excellent surface materials. It is also equipped with the convenient scene editor and C++ API for self-customized plug-in development.

Table \ref{tab:irs_composition} concludes the data distribution of IRS. IRS contains more than 100,000 pairs of 960x540 resolution stereo images (84,946 training and 15,079 testing), covering four indoor scene types and 70 different scene instances. The scenes are all enclosed indoor layout, and some of them even have a visible distance longer than 20 meters. We place over 2,091 identical objects of different types in the constructed space. We also consider different cases of brightness and light behaviors commonly happening in the indoor environments. 
\begin{table}[!ht]
	\centering
	\caption{The indoor scene types and visual attributes covered by our IRS.}
	\label{tab:irs_composition}
	\begin{tabular}{|c|c|} \hline
		Rendering Variable & Options \\ \hline\hline
		indoor scene & home(30995), office(41987), \\ &  restaurant(20969), store(6347) \\ \hline
		object & desk, chair, sofa, glass, mirror, \\ & 
		bed, bedside table, lamp, wardrobe, etc. \\ \hline
		brightness & over-exposure($>$1300), darkness($>$1700)\\ \hline
		light behavior & bloom($>$1700), lens flare($>$1700), \\   & glass transmission($>$3600), \\ 
		& mirror reflection($>$3600) \\ \hline
	\end{tabular}
\end{table}
\section{Quantitative Analysis}\label{sec:quan_analysis}
As supervised deep learning can only learn what they see in the dataset, the data characteristics can significantly affect the model accuracy and generalization. For the stereo matching problem, two of the most important factors are the disparity and the brightness, that shows the geometric information and light conditions respectively. In this section, we will quantitatively analyze the disparity distribution and brightness distribution of our IRS and compare it with some existing stereo datasets, including KITTI2012, KITTI2015, APOLLO, Sintel and FlyingThings. Besides, we also discuss the normal distribution which is important for the stereo matching, especially for the indoor scenes where there exist many texture-less regions, such as white wall and ceiling.

\subsection{Normal Analysis}
As IRS is the only stereo dataset that provides normal information among these datasets, we will only count the normal distribution of IRS. We first convert the three-dimensional normal vector $  n = (n_x,n_y,n_z) $ into two dimensional $ n_{angle} = ( \alpha = atan2(n_y,n_x) , \beta = atan2(n_z,\sqrt{n_x^2+n_y^2} ) $, where $ alpha $ represents the angle of $ n $ in the $x-y$ plane with the range $ [0^\circ,360^\circ] $, and $ beta $  represents the angle between $ n $ and the $x-y$ plane, ranging from $ [-90^\circ,0^\circ] $. Notice that $ beta \in [0^\circ,90^\circ] $  is invisible, because these regions are back toward the camera.
\begin{figure}[htbp]
	\centering
	\includegraphics[width=0.96\linewidth]{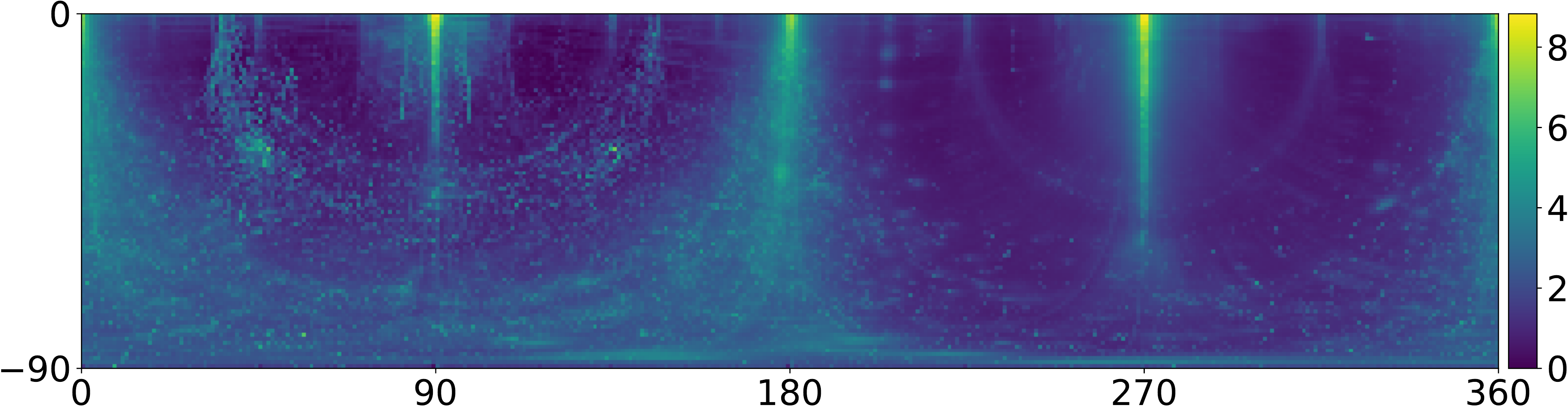}
	\caption{Normal Distribution}
	\label{fig:normal_distribution}
\end{figure}

We average the normal distributions in each sample of IRS and then transform the results by $log$ function for display purpose, as shown in Fig.  \ref{fig:normal_distribution}. As the images of IRS are captured indoors, the normal is mostly located around five directions, $ ( 0^\circ, 0^\circ) $, $ ( 90^\circ, 0^\circ) $, $ (180^\circ, 0^\circ) $, $ ( 270^\circ, 0^\circ) $ and $ (:, -90^\circ )$, which respectively correspond to the left wall, the floor, the right wall, the ceiling and the wall facing the camera. We hope these normal can provide useful information for the networks to estimate disparities in these texture-less regions.

\subsection{Disparity Analysis}
The main purpose of stereo matching is to find out the disparities between the left and the right, so it is important to have an insight of the disparity distribution of the dataset. Considering that different datasets have different image width, we normalize the disparity by image width using this formula $ disparity/width $ and then enlarge those divided results by 200$\times$ for better visualization.
\begin{figure}[htbp]
	\centering
	\subfloat[KITTI 2012]
	{
		\includegraphics[width=0.31\linewidth]{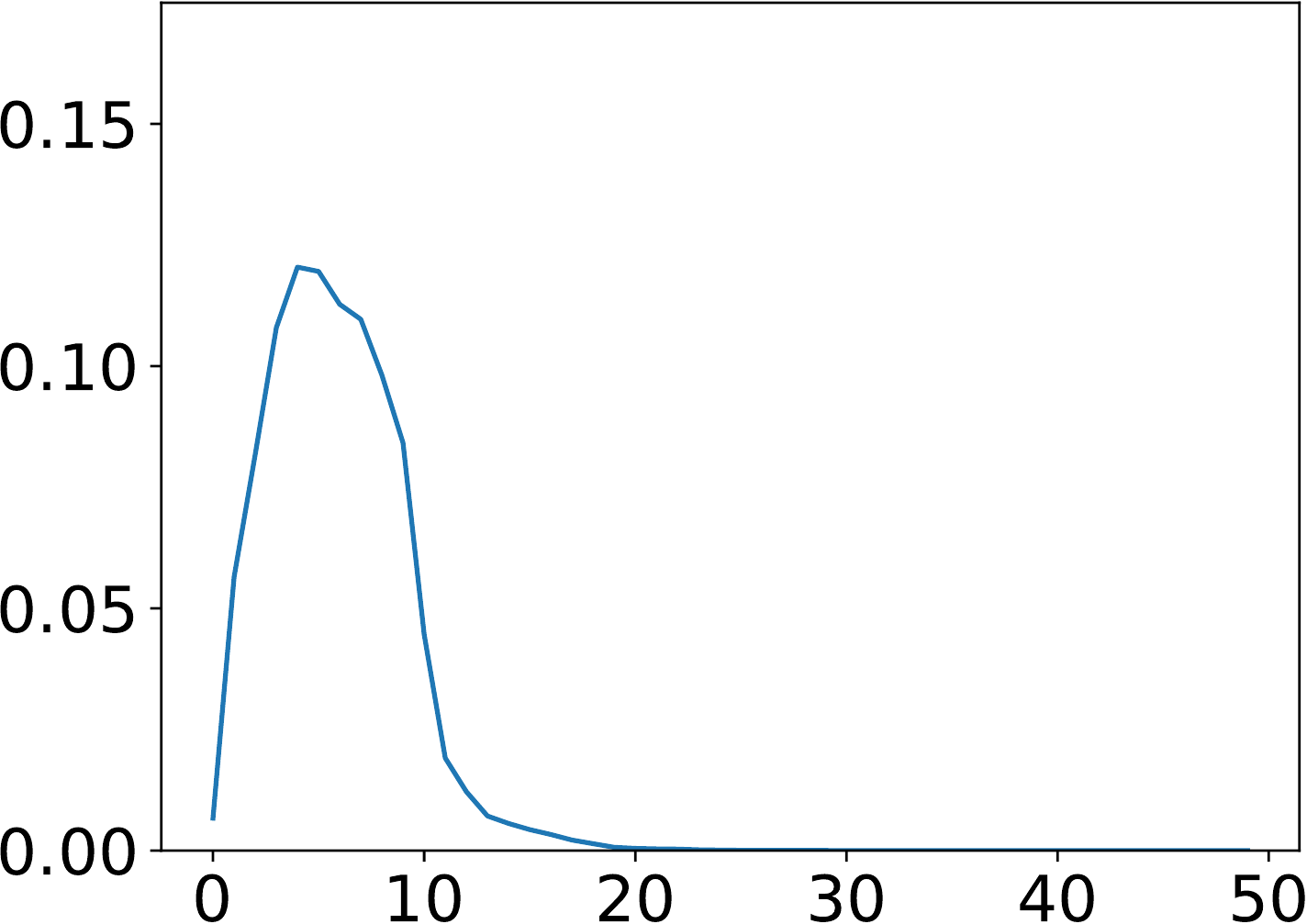}\label{fig:disp_kitti2012}
	}
	\subfloat[KITTI 2015]
	{
		\includegraphics[width=0.31\linewidth]{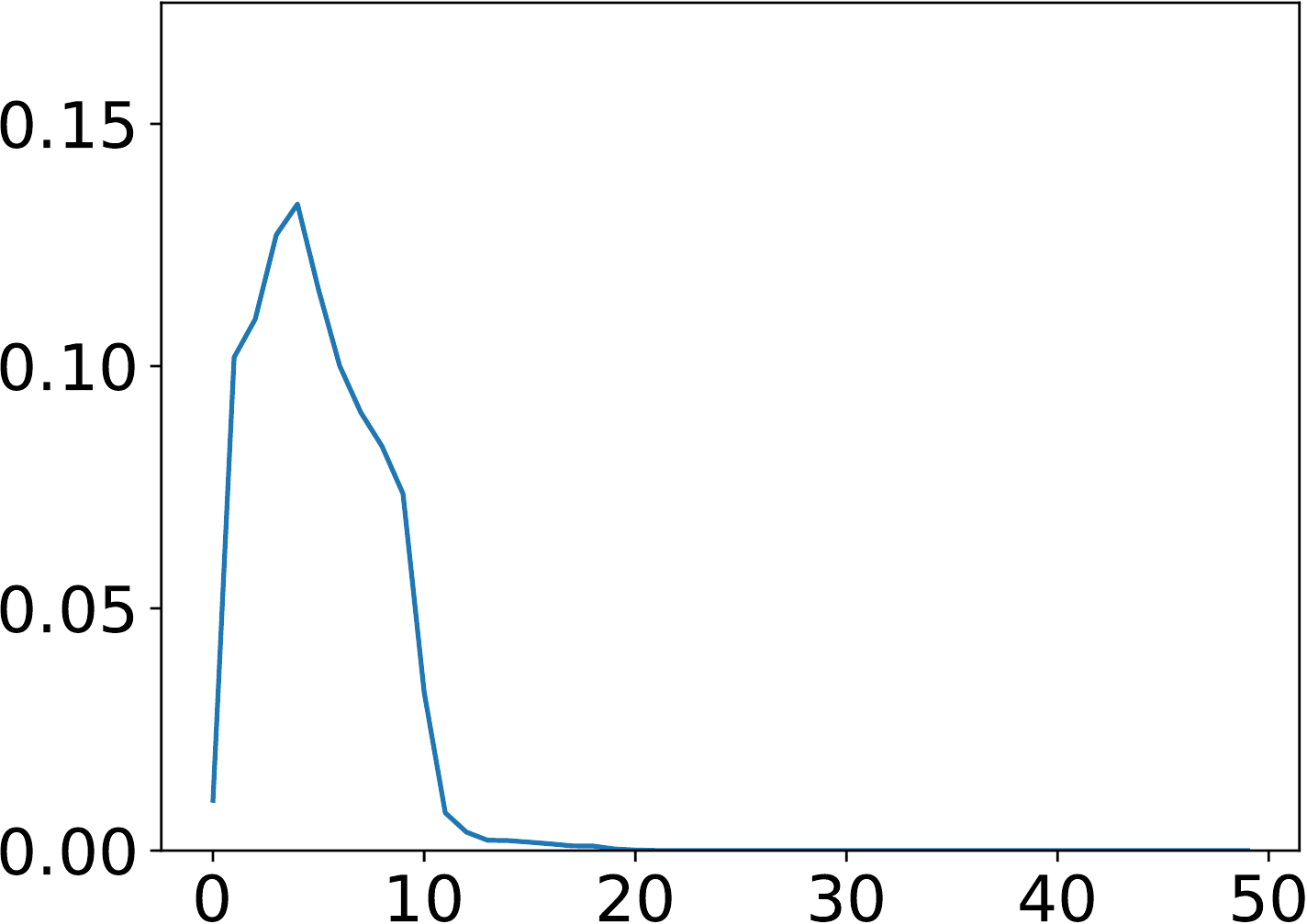}\label{fig:disp_kitti2015}
	}
	\subfloat[APOLLO]
	{
		\includegraphics[width=0.31\linewidth]{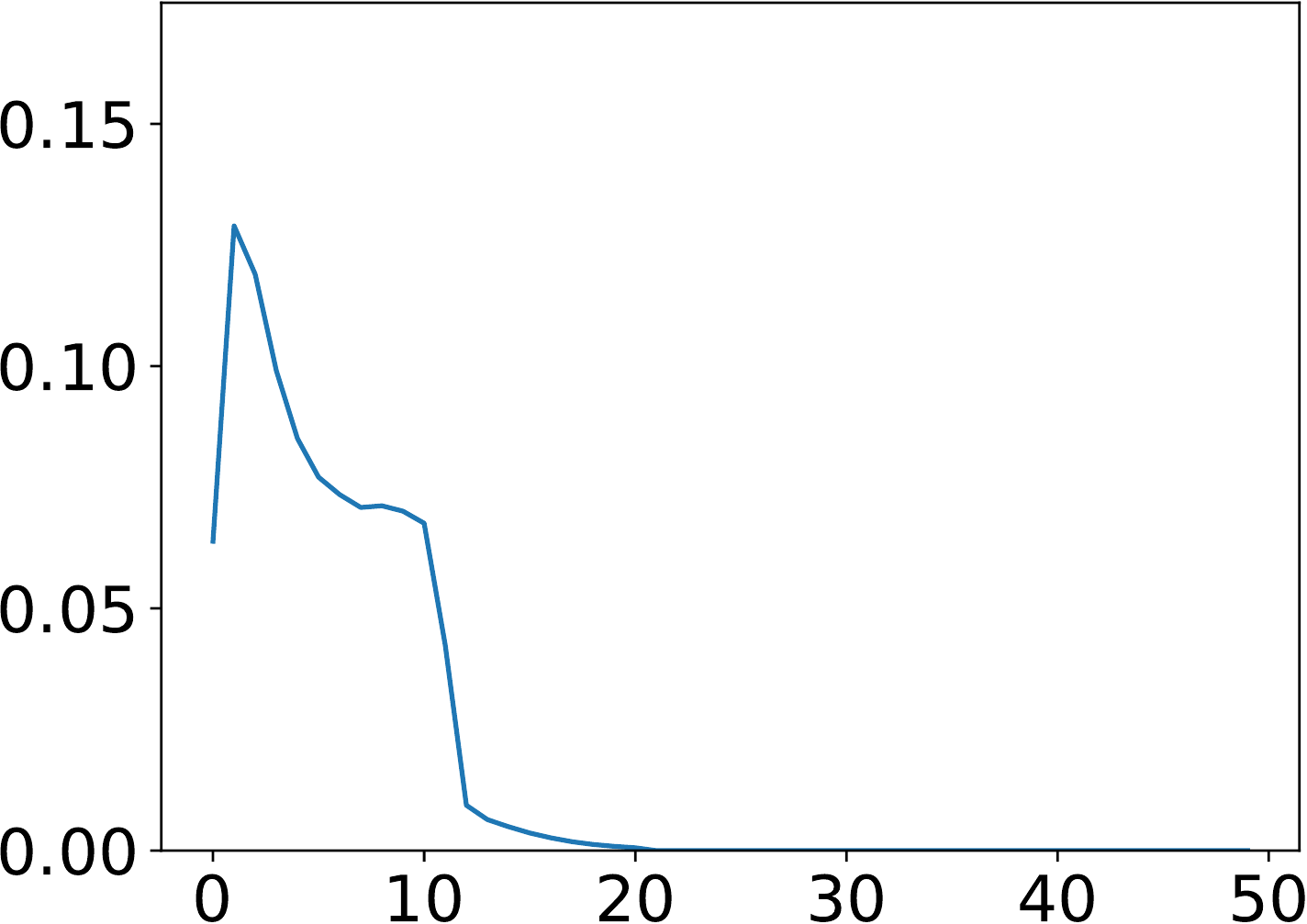}\label{fig:disp_apollo}
	}
	\qquad
	\vspace{-0.5 em}
	\subfloat[SINTEL]
	{
		\includegraphics[width=0.31\linewidth]{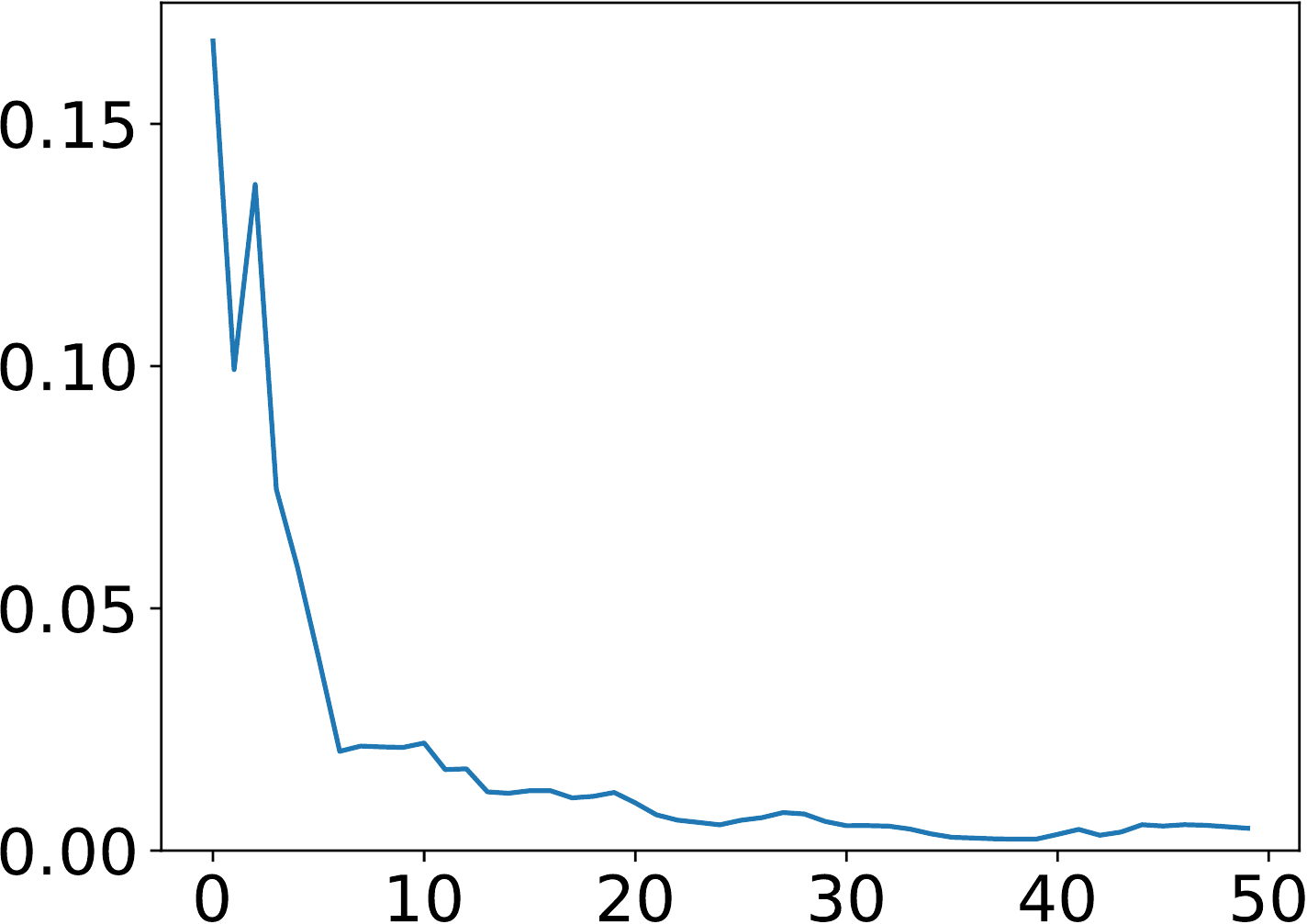}\label{fig:disp_sintel}
	}
	\subfloat[FLYINGTHINGS]
	{
		\includegraphics[width=0.31\linewidth]{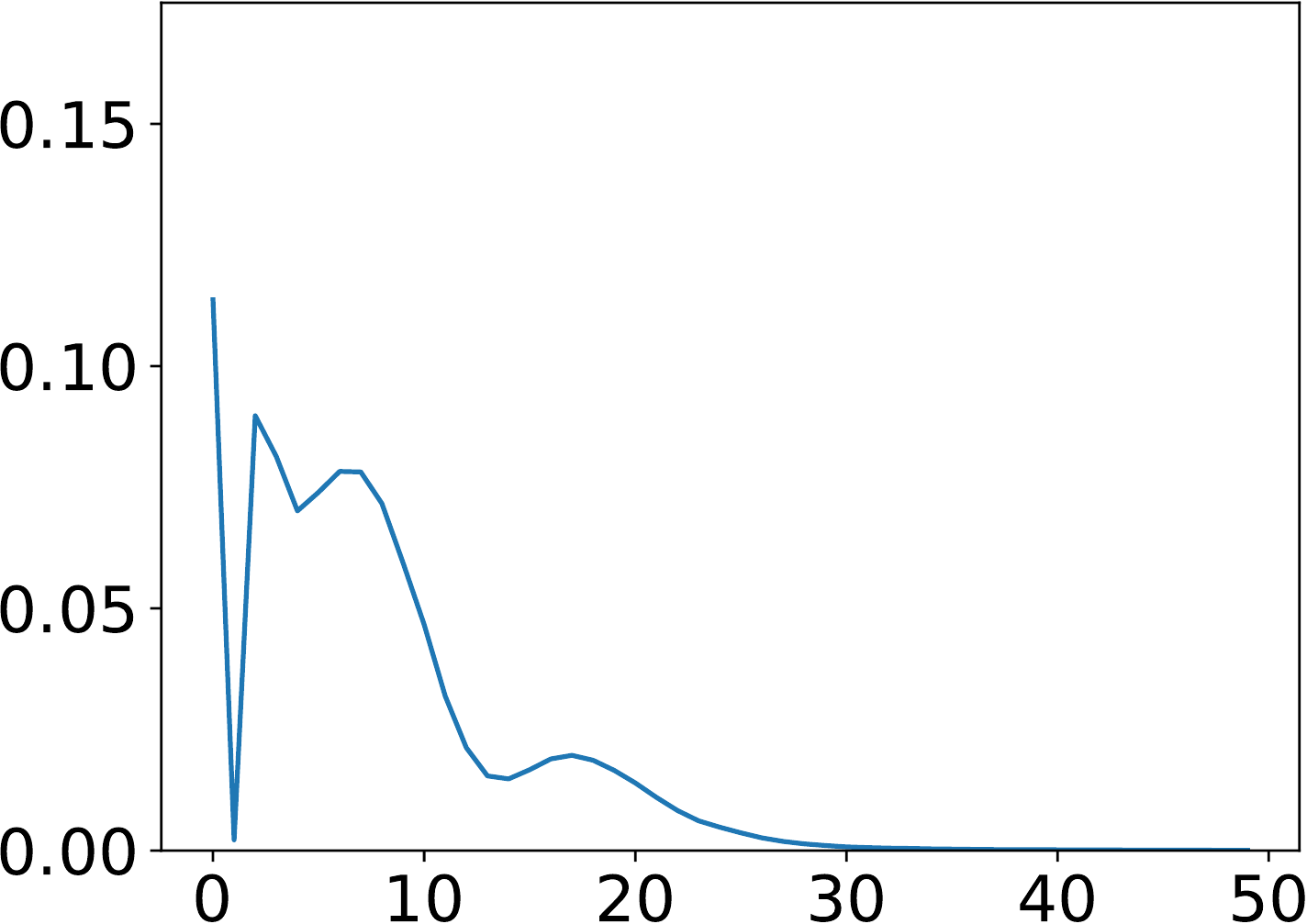}\label{fig:disp_flying}
	}
	\subfloat[OURS]
	{
		\includegraphics[width=0.31\linewidth]{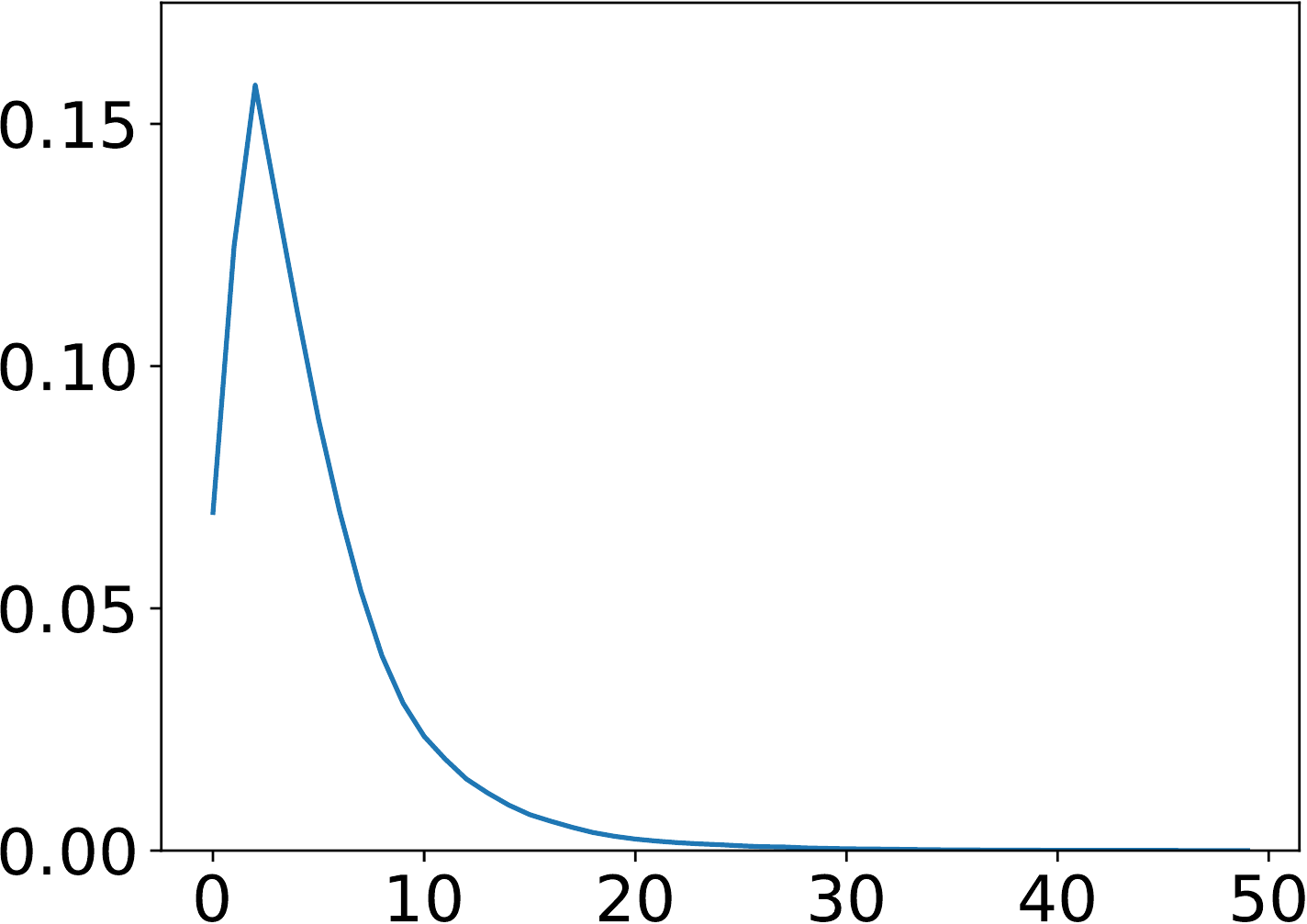}\label{fig:disp_ours}
	}
	\caption{Disparity Distribution.}
	\label{fig:disparity_distribution}
\end{figure}

We draw the disparity distribution of six different datasets in Fig. \ref{fig:disparity_distribution}, where the $x$ axis represents the preprocessed disparity values and the $y$ axis represents the percentage. As most of the values are less than 50, we only present the range of $ [0,50] $. In Fig. \ref{fig:disparity_distribution}, (a)-(c) are the natural datasets and (d)-(f) are the synthetic datasets. Sintel and Scene Flow are two of the most commonly used datasets, but there are some obvious troughs in their distributions, which are quite different from the natural datasets and may lead to wrong results when we apply models trained on these dataset to natural environment. However, our proposed IRS has no such problem and exhibits a similar distribution to those natural datasets. Besides, we also find that there is some difference from $ 5\% $ to $ 10 \% $ between the APOLLO, the KITTI2012, the KITTI2015 and the IRS. The reason behind this phenomenon is that the IRS is captured indoors and the rest datasets are captured outdoors, which explains why the model trained on indoor dataset cannot be used in outdoor dataset and vice versa. This difference also shows the importance and concern of the IRS, which provides a large set of indoor stereo images and focuses on the indoor robotics applications. 

\subsection{Brightness Analysis}
Besides the disparity distribution, another key factor of dataset is the brightness distribution, which reflects the light condition of the environment. It is more important for the indoor environment than the outdoor environment, because there may be many lighting sources in indoor, such as the sunlight from the windows and different kinds of light bulbs, whilst the main light source for outdoor is the sun. These factors make it easier for the camera to overexpose in indoor environment. Considering this, we try to analyze the brightness distribution and show the advantage of the IRS.

\begin{figure}[htbp]
	\centering
	\subfloat[KITTI 2012]
	{
		\includegraphics[width=0.3\linewidth]{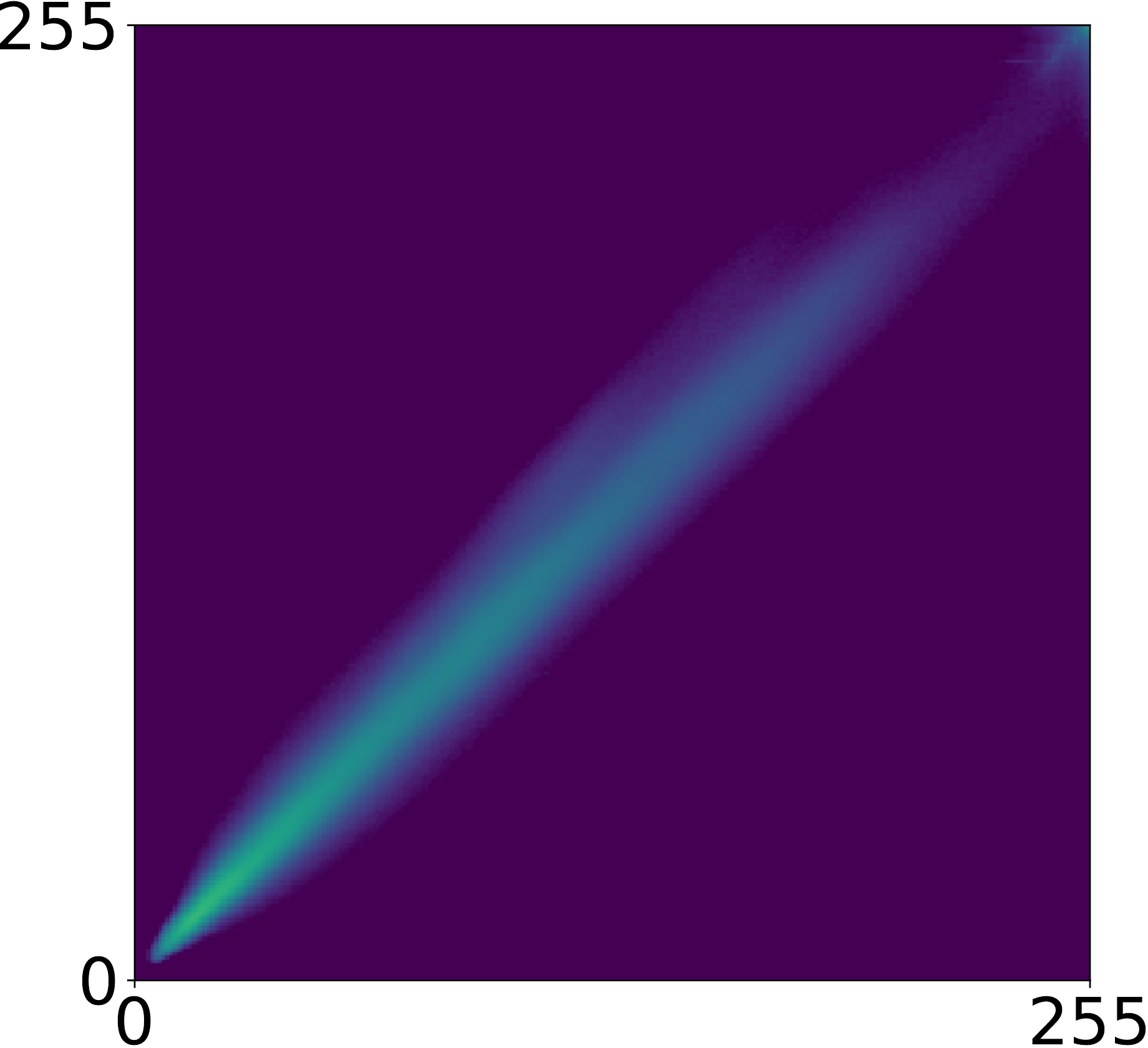}\label{fig:lumin_kitti2012}
	}
	\subfloat[KITTI 2015]
	{
		\includegraphics[width=0.3\linewidth]{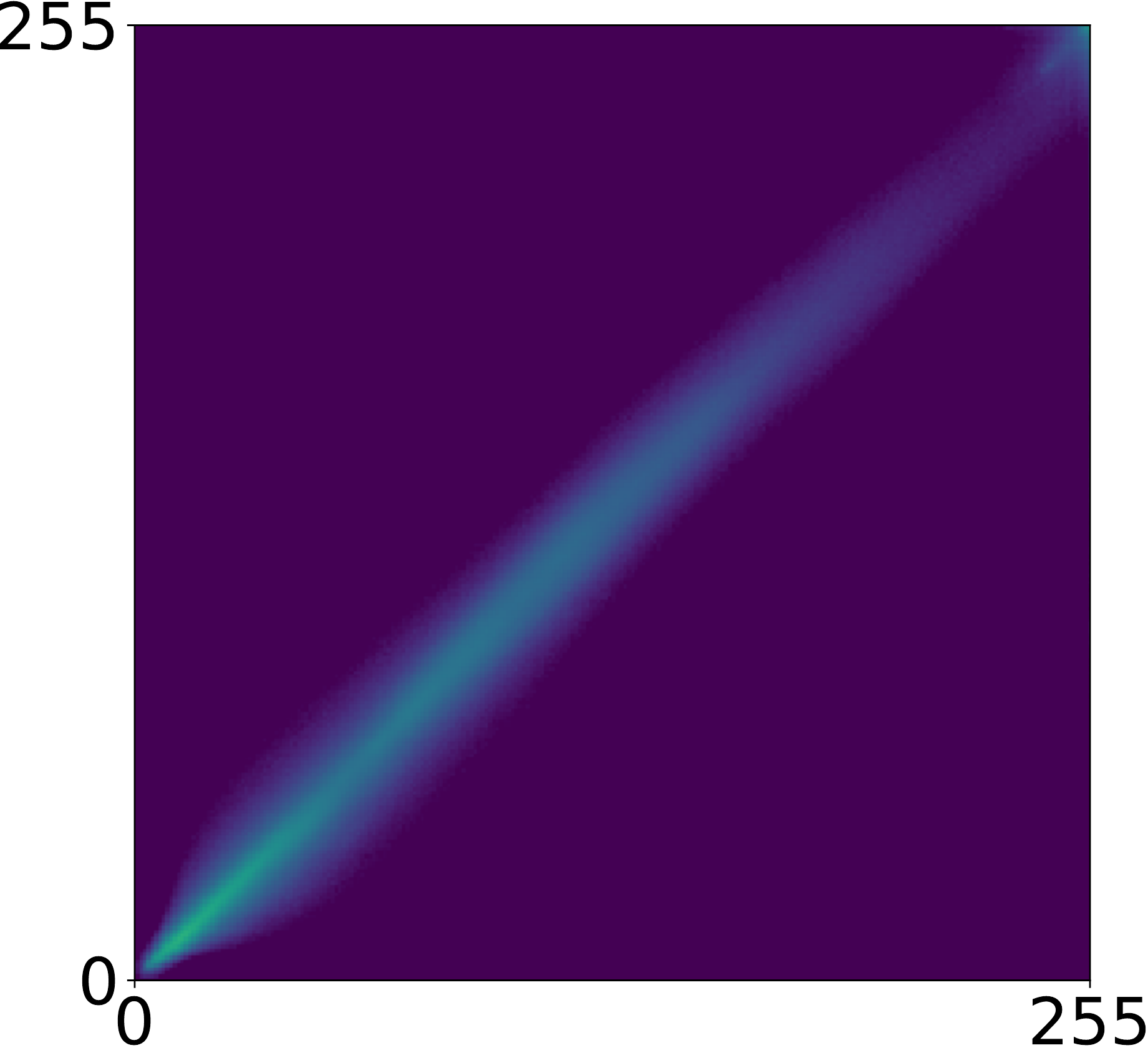}\label{fig:lumin_kitti2015}
	}
	\subfloat[APOLLO]
	{
		\includegraphics[width=0.335\linewidth]{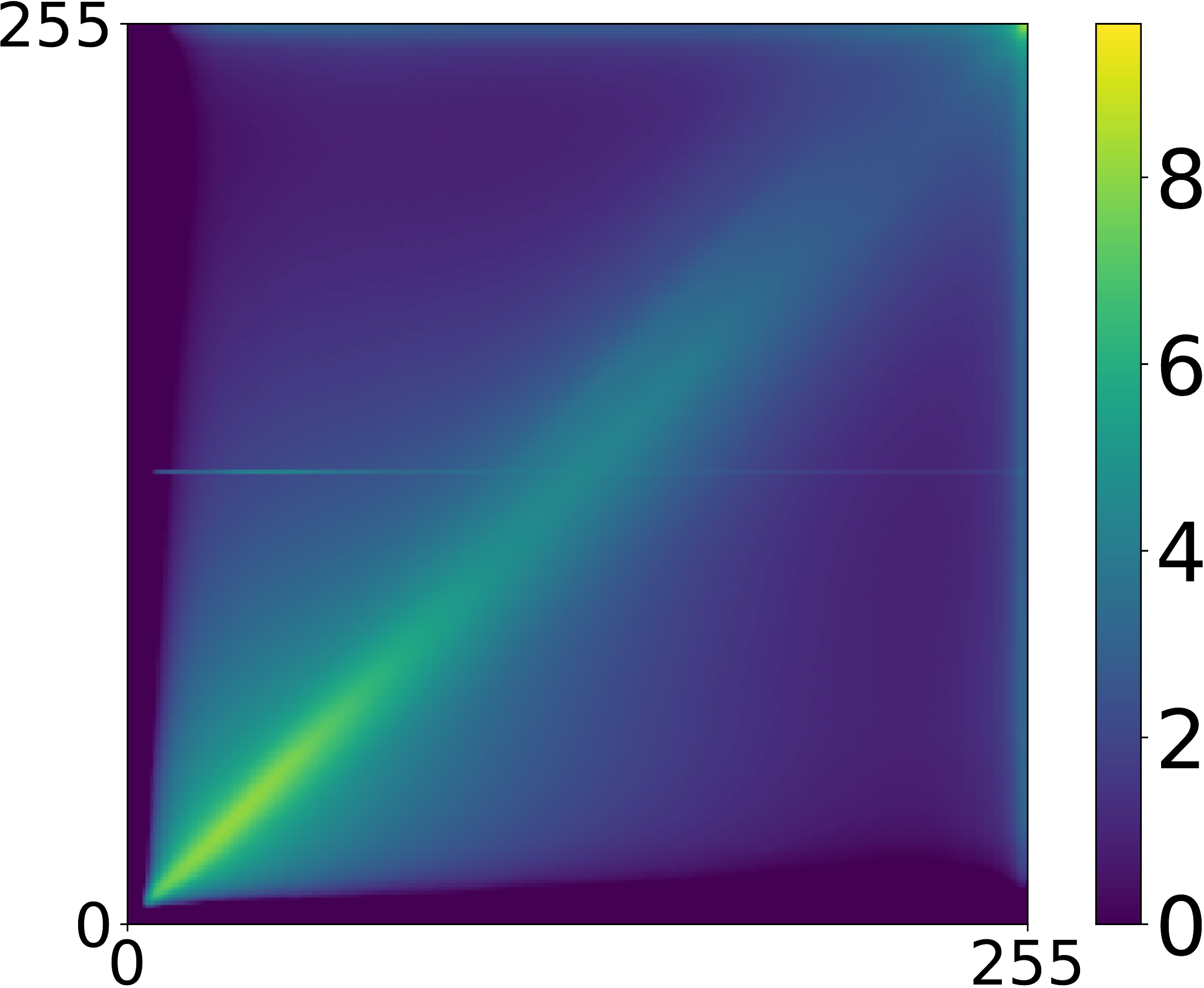}\label{fig:lumin_apollo}
	}
	\qquad
	\subfloat[SINTEL]
	{
		\includegraphics[width=0.3\linewidth]{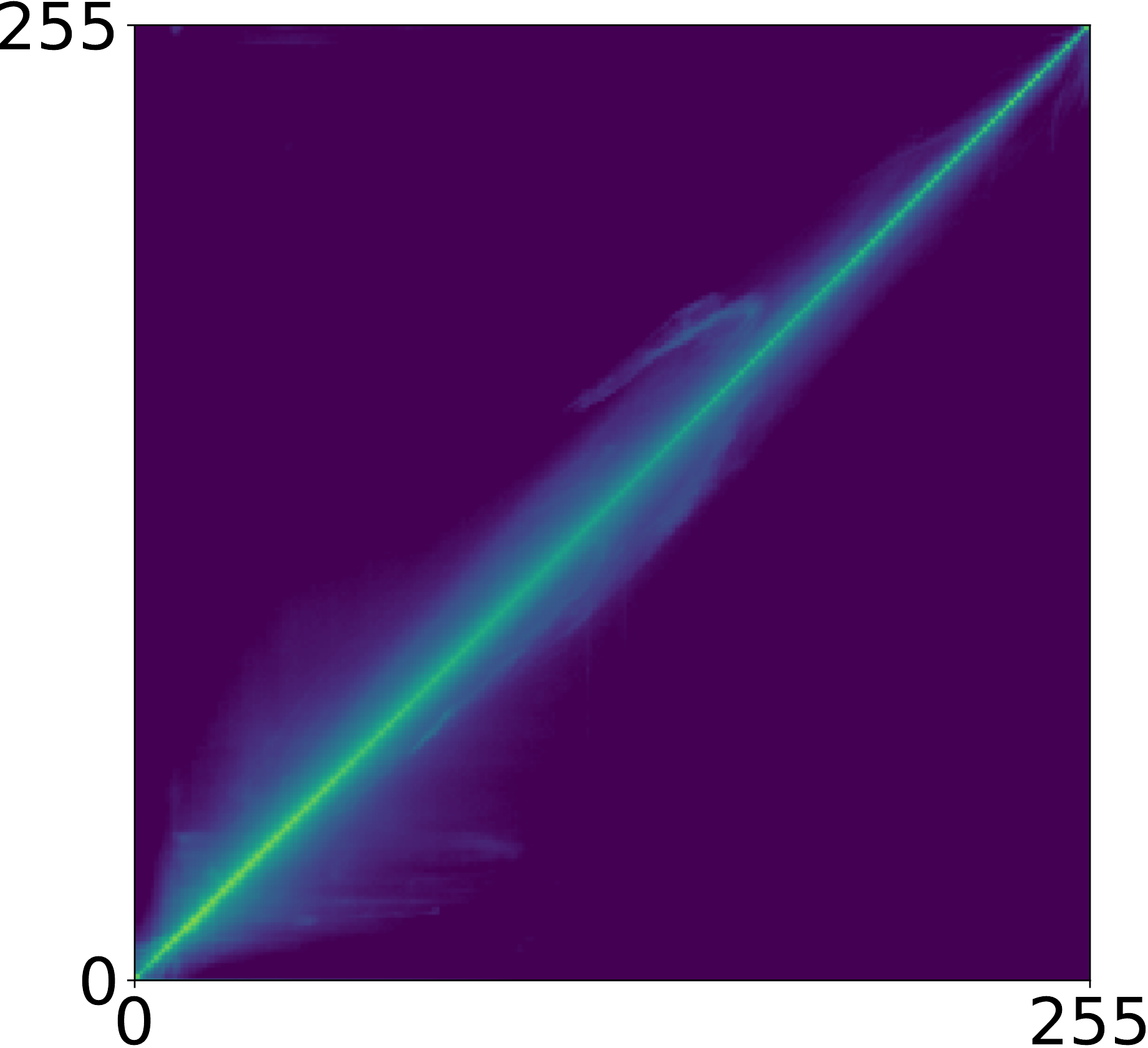}\label{fig:lumin_sintel}
	}
	\subfloat[FLYINGTHINGS]
	{
		\includegraphics[width=0.3\linewidth]{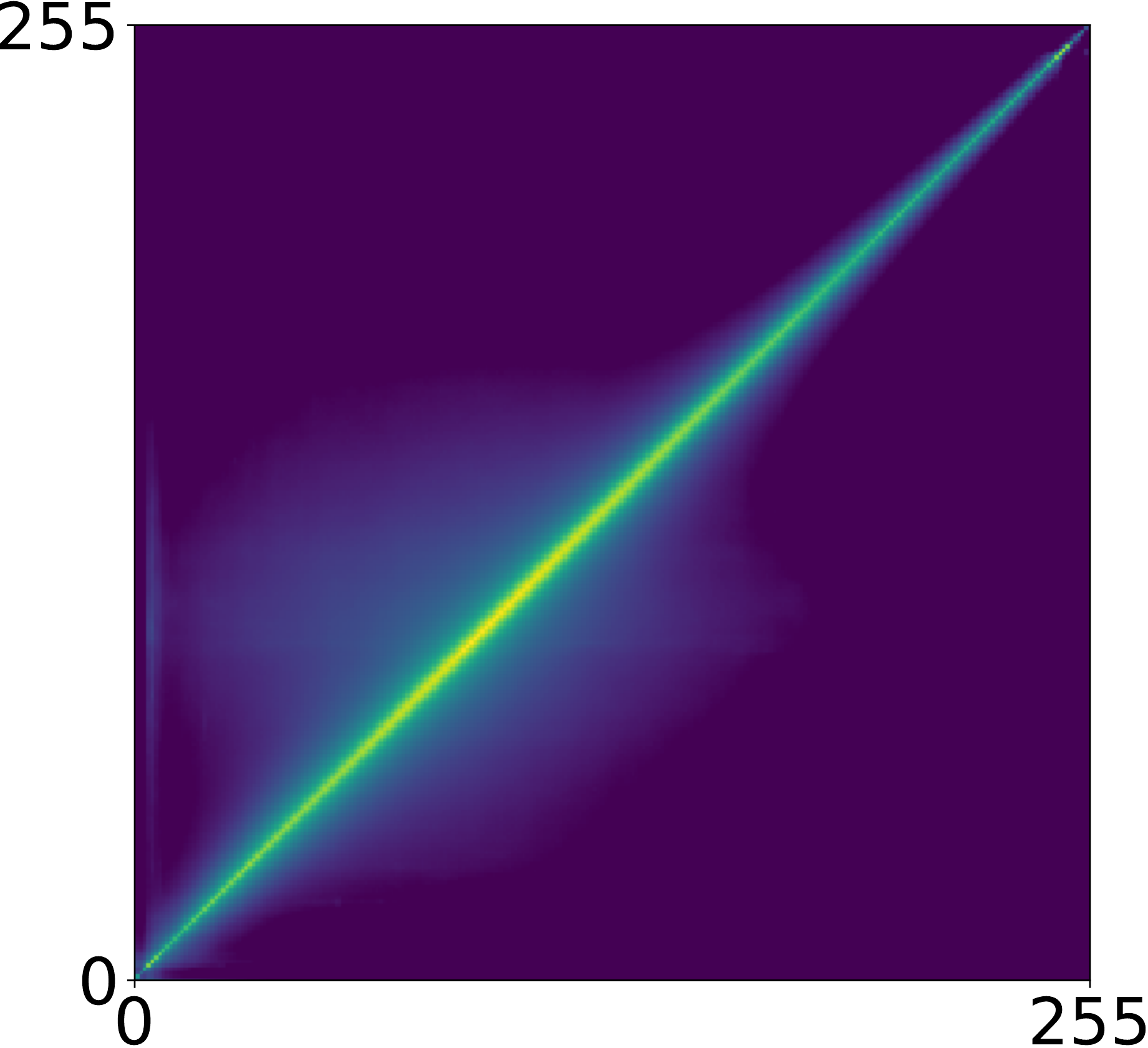}\label{fig:lumin_flying}
	}
	\subfloat[OURS]
	{
		\includegraphics[width=0.335\linewidth]{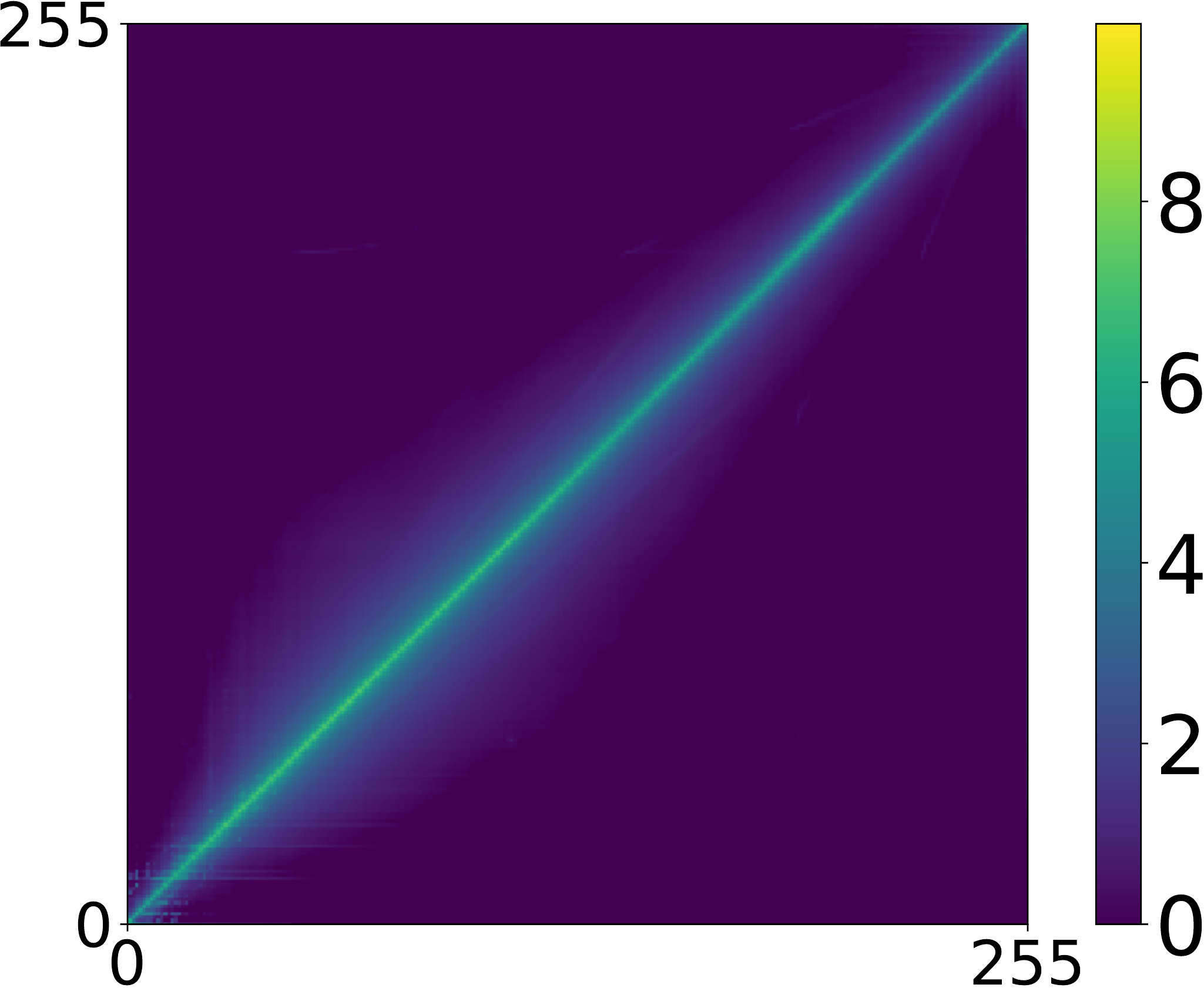}\label{fig:lumin_ours}
	}
	\caption{Brightness Distribution.}
	\label{fig:luminanace_distribution}
\end{figure}

We first convert all the RGB images into gray-scale, using this formula $ 0.299R+0.587G+0.114B $. The results are shown in Fig. \ref{fig:luminanace_distribution}, where the $x$ and $y$ axes are the brightness of the left and right image respectively, and the value represents the logarithm of the average number of matched pixels on each pair. 

The distribution of matched pixels in each dataset roughly spreads along a line ($ y=x $). For three natural datasets, KITTI 2012, KITTI 2015 and APOLLO, the distribution is relatively discrete due to the uncontrollable external environments, which is caused by the multiple diffraction of the light. The distinctive distribution pattern of APOLLO is caused by different weather and climate conditions. However, in the synthetic datasets, matched pixels are much closer to $y=x$, which is caused by the limitation of ray tracing. The distribution patterns of IRS and SINTEL are similar, while FLYINGTHINGS has volatilization in the low-value range. The reason is that they use different 3D animation engines. Data enhancement is needed to fill this gap between natural dataset and synthetic dataset.

\begin{figure}[htbp]
	\centering
	\subfloat[KITTI 2012]
	{
		\includegraphics[width=0.3\linewidth]{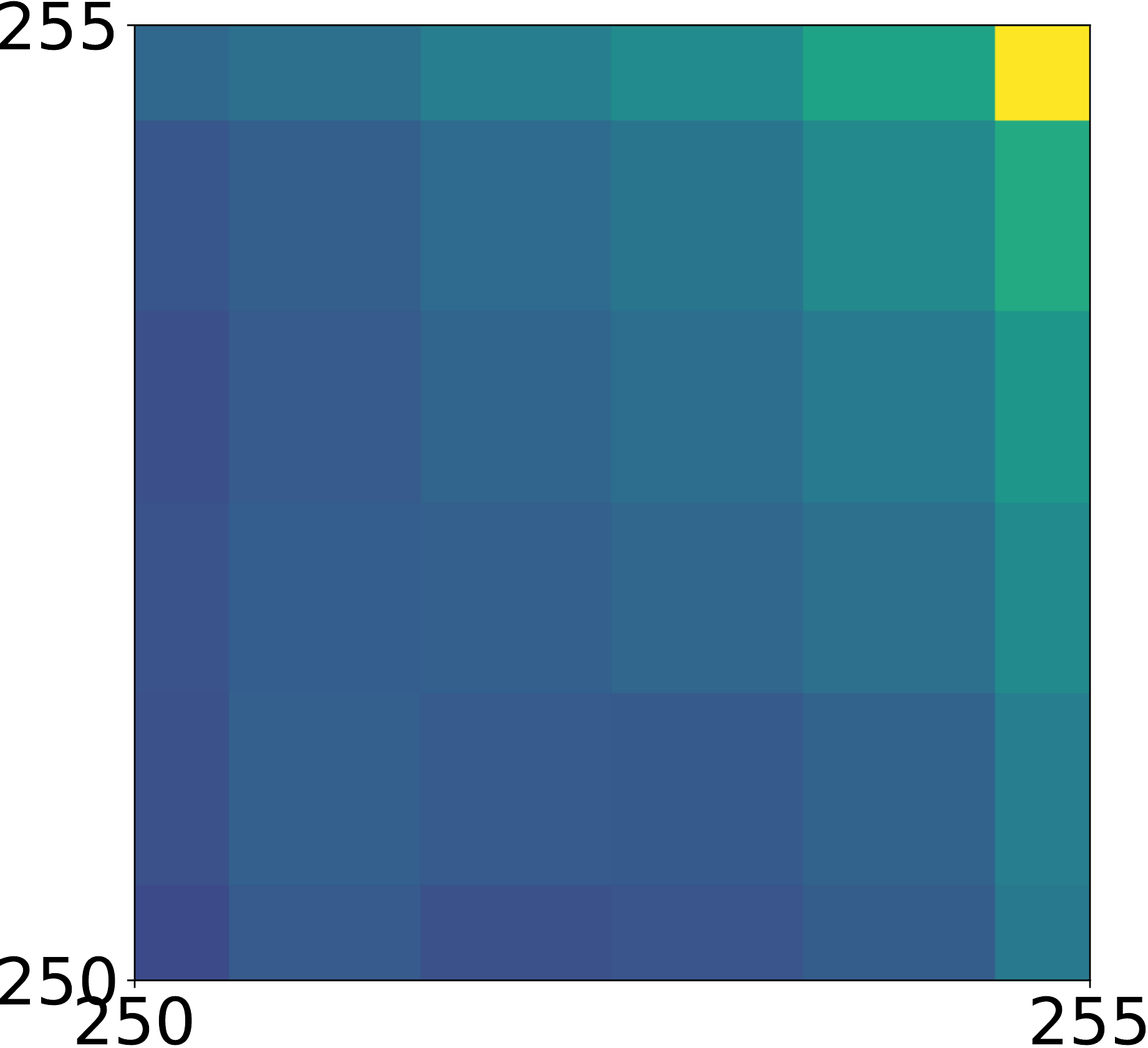}\label{fig:part_lumin_kitti2012}
	}
	\subfloat[KITTI 2015]
	{
		\includegraphics[width=0.3\linewidth]{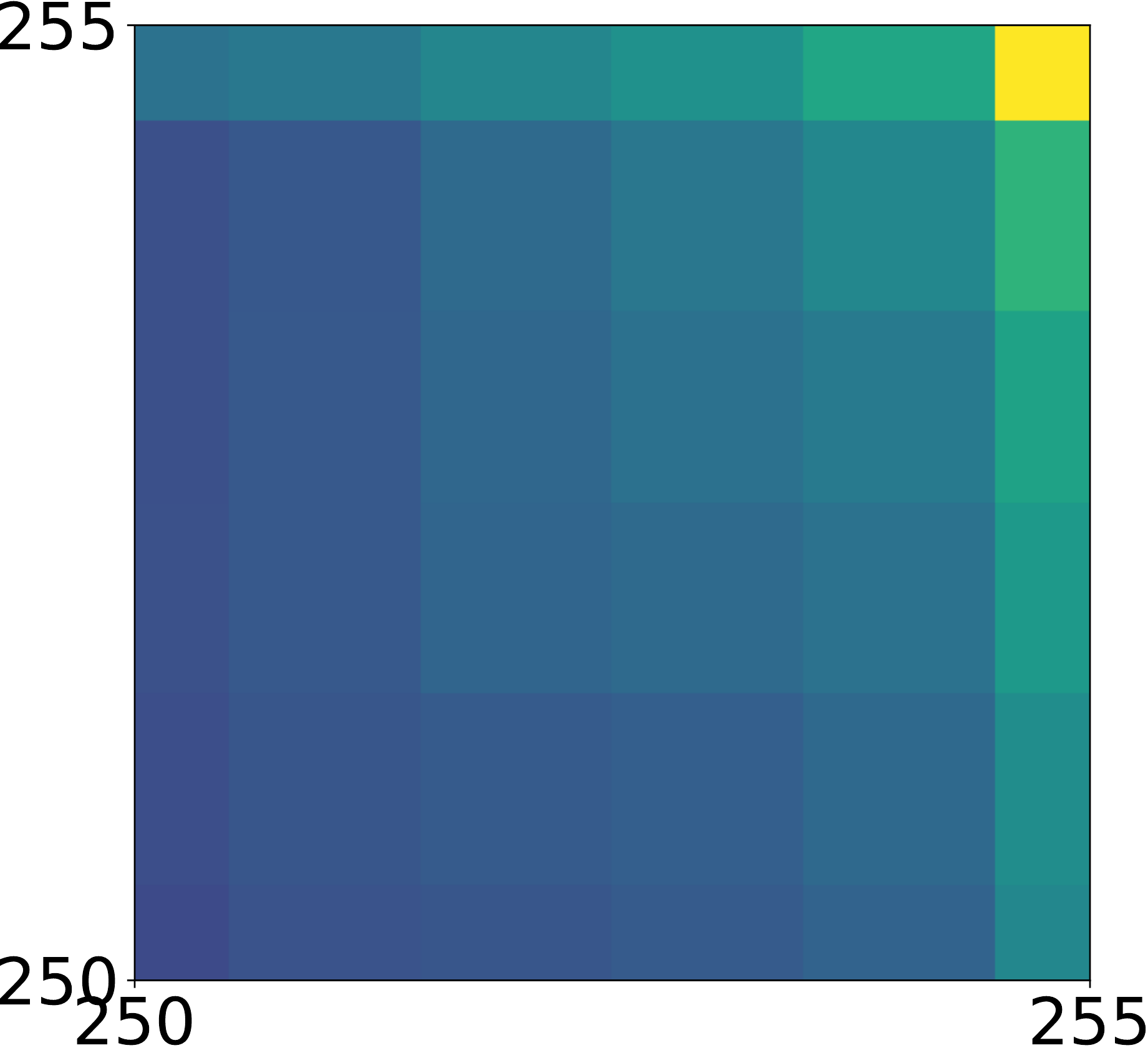}\label{fig:part_lumin_kitti2015}
	}
	\subfloat[APOLLO]
	{
		\includegraphics[width=0.335\linewidth]{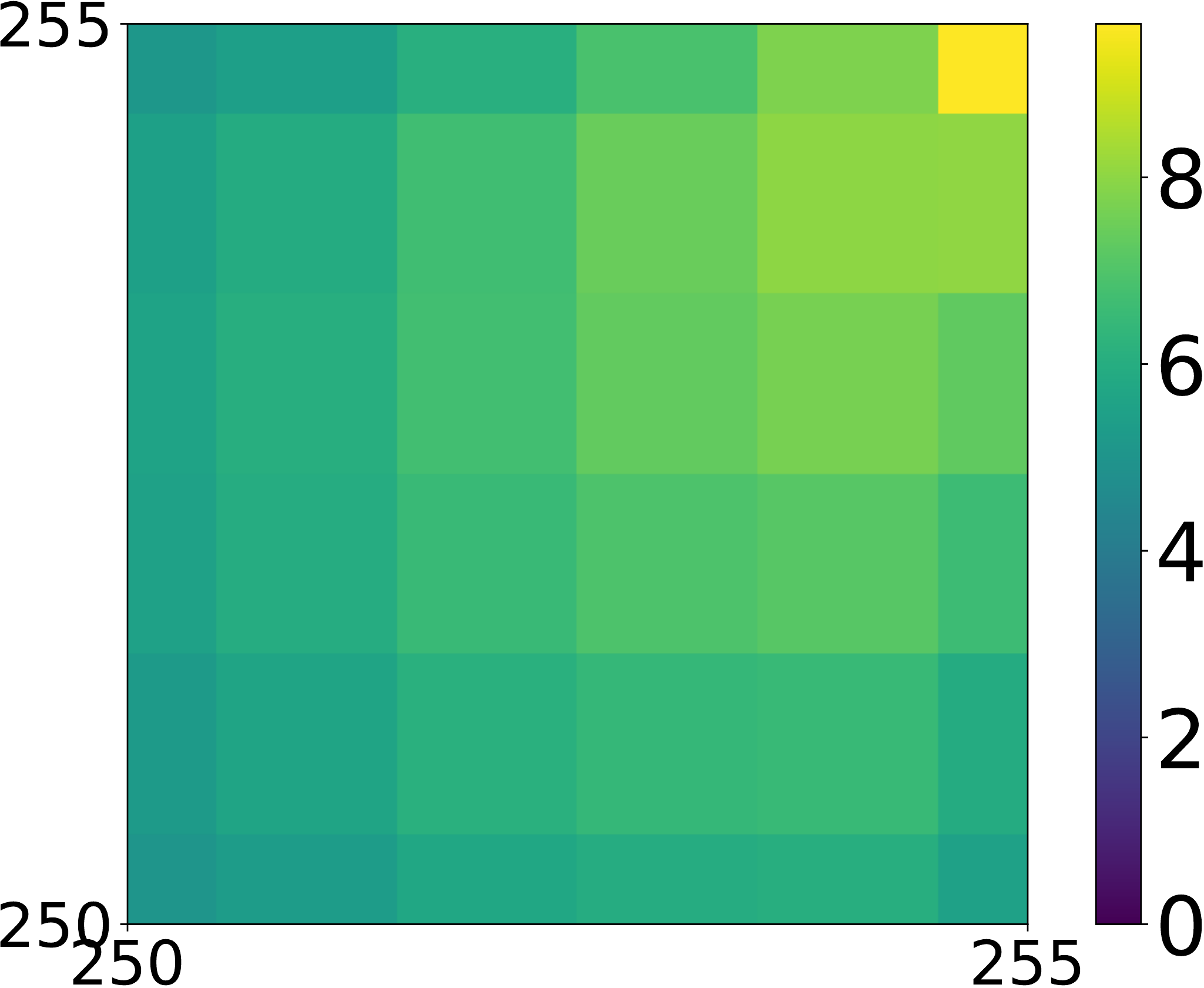}\label{fig:part_lumin_apollo}
	}
	\qquad
	\subfloat[SINTEL]
	{
		\includegraphics[width=0.3\linewidth]{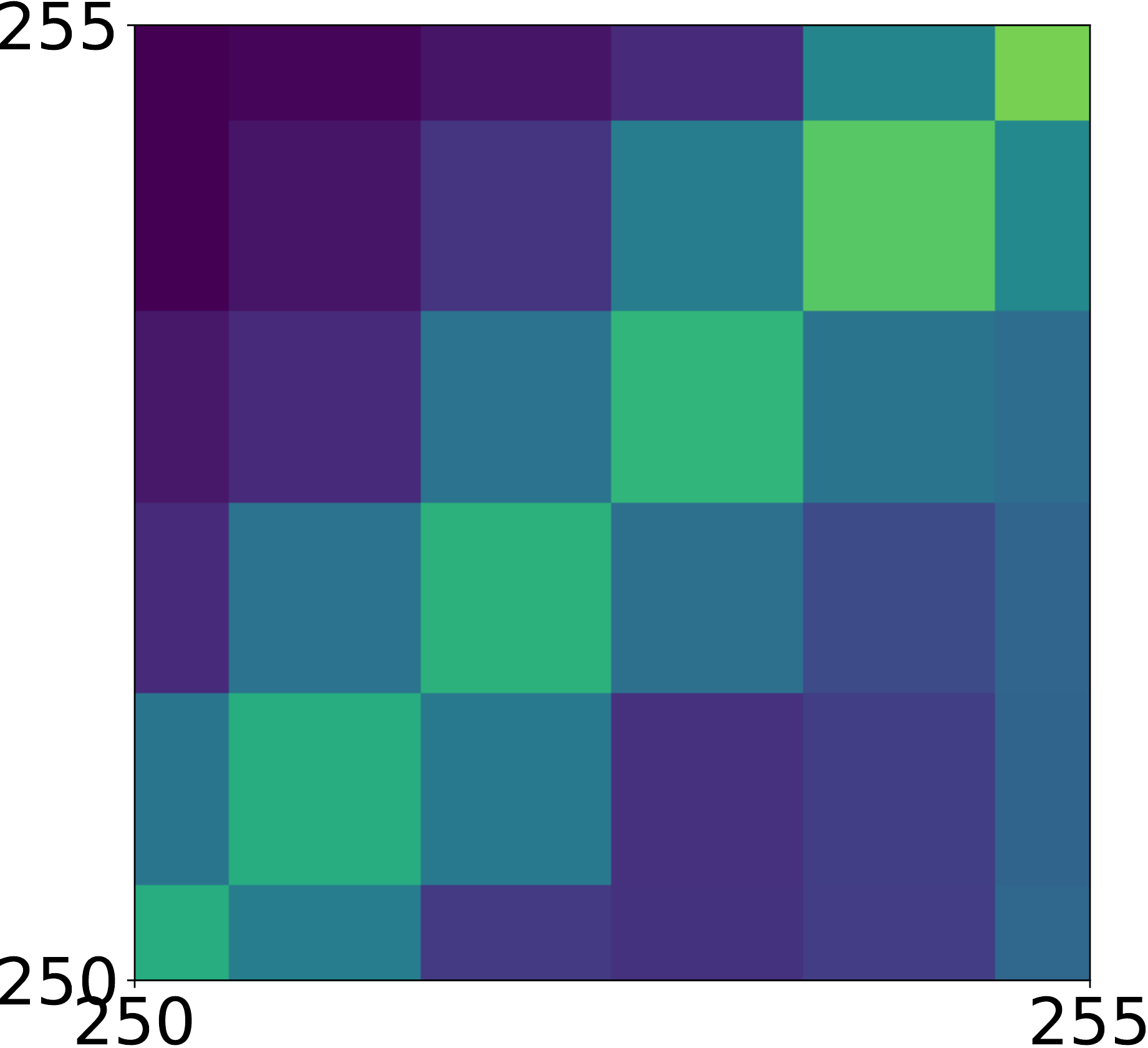}\label{fig:part_lumin_sintel}
	}
	\subfloat[FLYINGTHINGS]
	{
		\includegraphics[width=0.3\linewidth]{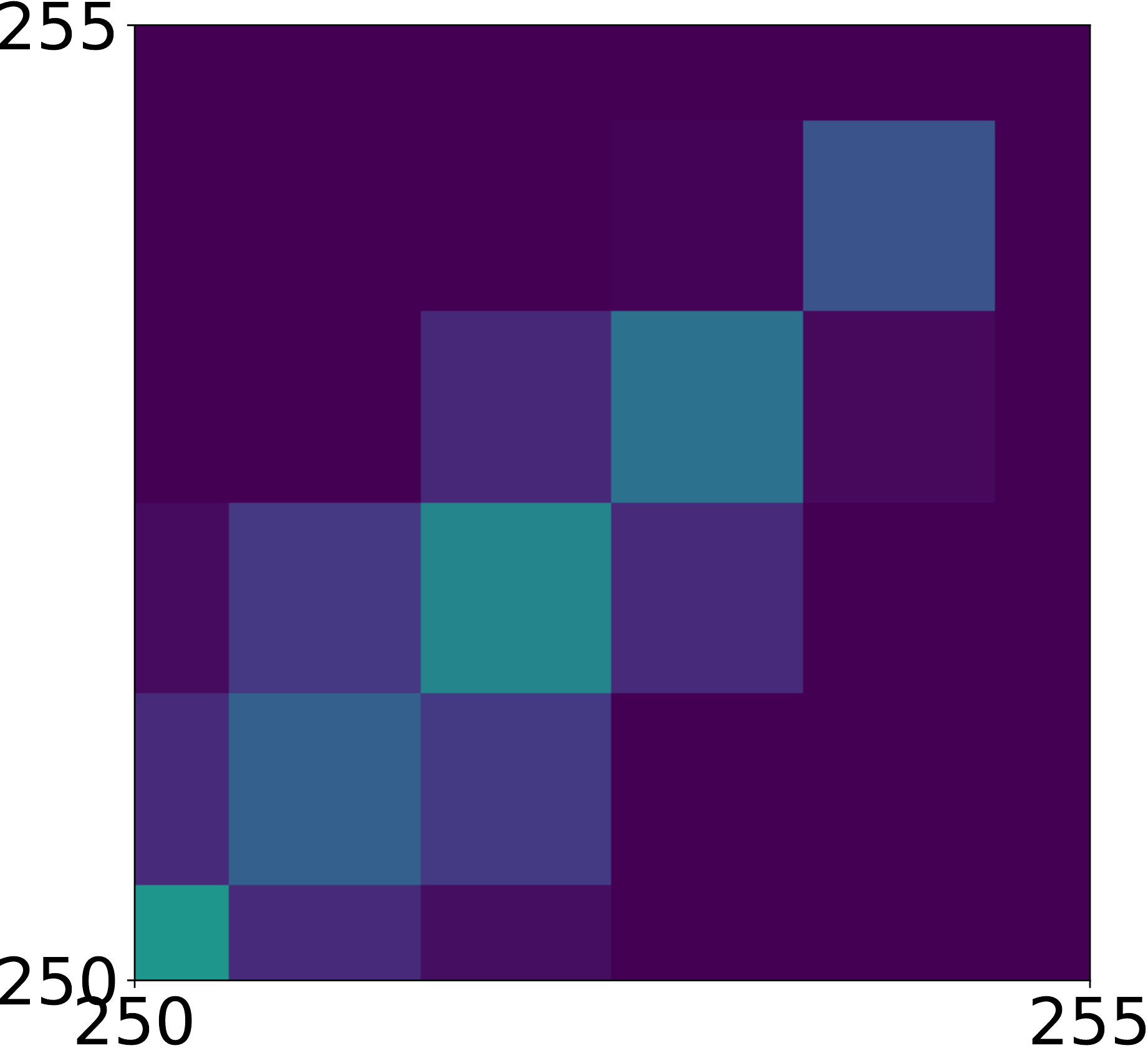}\label{fig:part_lumin_flying}
	}
	\subfloat[OURS]
	{
		\includegraphics[width=0.335\linewidth]{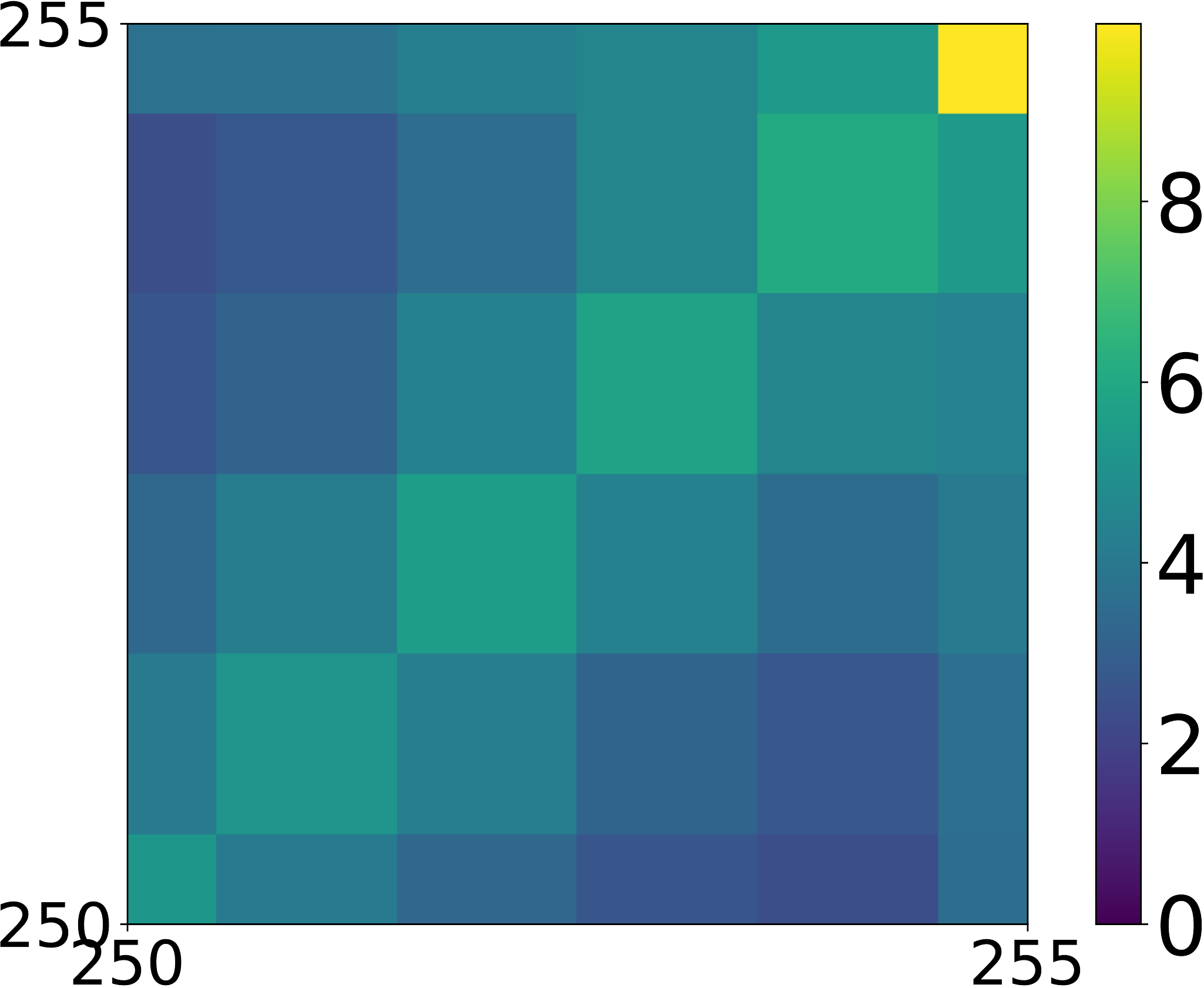}\label{fig:part_lumin_ours}
	}
	\caption{Over-Exposure Pixels Distribution.}
	\label{fig:part_lumin_distribution}
\end{figure}

In addition, in the natural datasets, overexposure is a common phenomenon, which is caused by the changing light environment, the limitation of digital camera and so on. In Fig. \ref{fig:part_lumin_distribution}, the three natural datasets, the KITTI 2012, the KITTI 2015 and the Apollo, have a large number of matched overexposed pixels at $ (255,255) $; but the existing synthetic datasets, like SINTEL and Scene Flow, do not have enough overexposure pixels. In IRS, we purposely create enough overexposed pixels to simulate the real scenarios, aiming at minimizing the gap.

\section{Networks}
To prove the applicability of our proposed synthetic datasets to disparity and surface normal estimation, we use them to train several deep models, including the existing state-of-the-art architectures and our own proposed one. 

\subsection{Disparity Estimation}
To validate the effectiveness of IRS for indoor disparity estimation, we used different training datasets to train the same DNN architecture for constructing different DNN models. Then we applied those models on different testing datasets and obtained the errors. We used the endpoint error (EPE) as error measure in all the cases. 

We chose FADNet \cite{wang2020fadnet} and GwcNet \cite{groupnet} from existing state-of-the-art networks, as they perform well in both accuracy and speed. For each network, we trained three models, respectively with IRS, FT3D (short for FlyingThings3D \cite{mayer2016large}) and F+I (short for both). Then we applied them to different testing datasets, including five synthetic ones listed in Table \ref{tab:epe_synthetic} and three real-world ones listed in Table \ref{tab:epe_real}.

\subsection{Normal Estimation}
\begin{figure}[ht]
	\centering
	\includegraphics[width=0.96\linewidth]{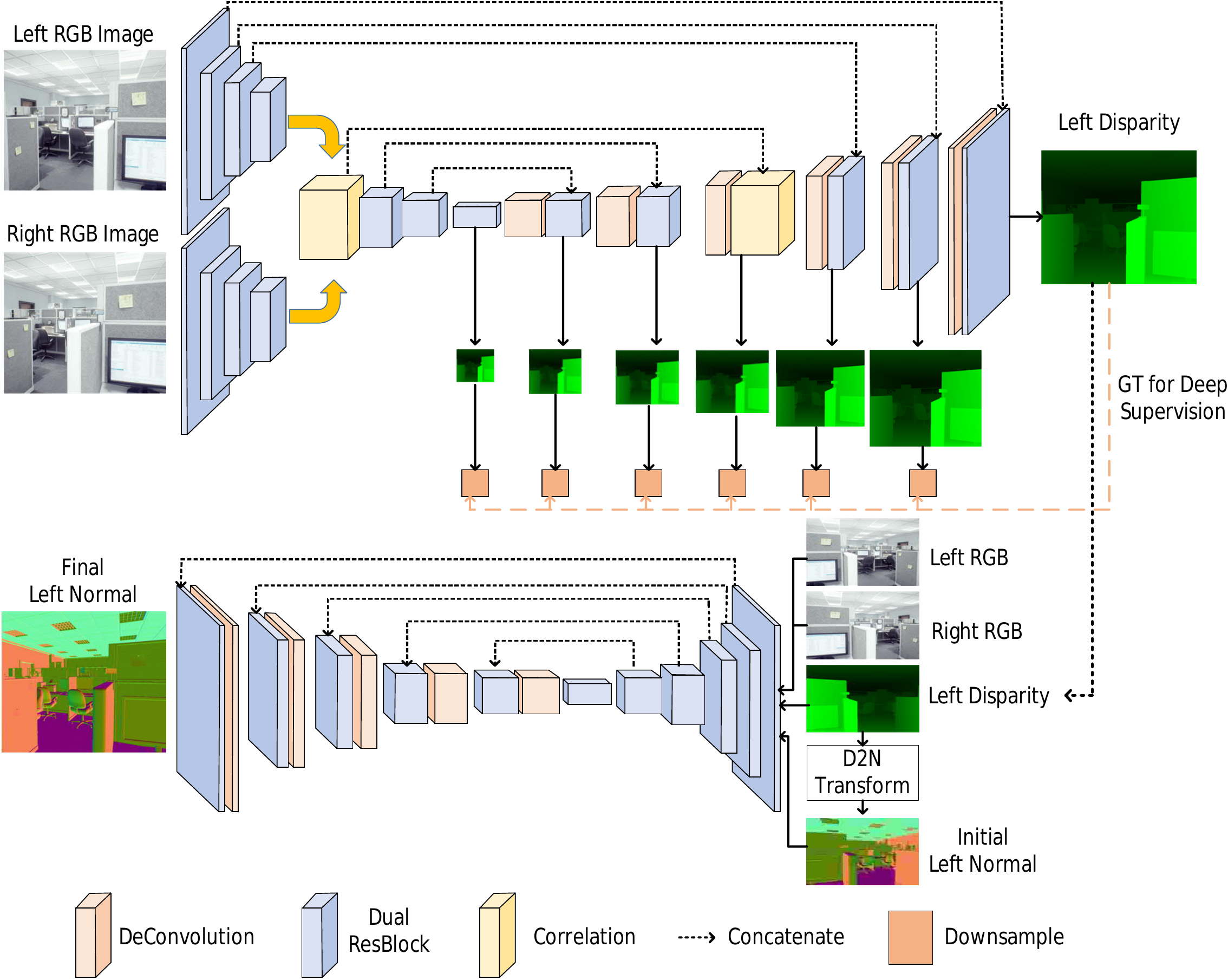}
	\caption{The Structures of Our proposed DTN-Net.}
	\label{fig:dtn-net}
\end{figure}

We designed a novel deep model, DTN-Net, to predict the surface normal map by refining the initial one transformed from the predicted disparity. Fig. \ref{fig:dtn-net} demonstrates the structure of DTN-Net (\underline{D}isparity \underline{T}o \underline{N}ormal Network), which is comprised of two modules, RD-Net \cite{wang2020fadnet} and NormNetS. First, RD-Net predicts the disparity map for the input stereo images. Then we apply the transformation from disparity to normal in \cite{geonet}, denoted by D2N Transform, to produces the initial coarse normal map. Finally, NormNetS takes the stereo images, the predicted disparity map by RD-Net and the initial normal map as input and predicts the final normal map. The structure of NormNetS is similar to DispNetS \cite{mayer2016large} except that the final convolution layer outputs three channels instead of one, as each pixel normal has three dimension $(x,y,z)$. As comparison, we also apply NormNetS with only the RGB stereo input, and DFN-Net(\underline{D}epth \underline{F}usion \underline{N}ormal Network), a RGB-D based scheme originally derived in \cite{Zeng_2019_CVPR}.

To compare the predicted normal to the ground truth, we first calculate the angle between them and take the mean value and the median value. We then compute the fraction of pixels of which the angle error is less than $t$, where $t=11.25^\circ, 22.5^\circ, 30^\circ$, as adopted in \cite{Zeng_2019_CVPR}.

\subsection{Training Implementation}
We implemented all the models using PyTorch. As for FADNet and GwcNet, we follow the original training settings and hyper-parameters in \cite{wang2020fadnet} and \cite{groupnet} respectively. As for DTN-Net, we basically follow the training scheme of training FADNet \cite{wang2020fadnet}, such as color normalization and cropping for input images, and multi-scale weight scheduling training. 

Due to the space limitation, we only highlight some significant findings in the experimental results. We detail the network structure, the training process and the complete experimental analysis in the supplementary material. 

\section{Experimental Results}\label{sec:exp}
\subsection{Benchmark Results for Disparity Estimation}
We explore the accuracy of FADNet and GwcNet trained by FlyingThings3D and IRS respectively on four synthetic stereo datasets in Table \ref{tab:epe_synthetic}. Take FADNet as an example. First, it is reasonable that FADNet(FT3D) and FADNet(IRS) perform better than each other on their own testing samples due to the similar RGB data distribution. However, FADNet(FT3D) has a high average EPE of 6.01 on IRS, while FADNet(IRS) still performs good generalization with a 2.36 average EPE on FlyingThings3D. Furthermore, merging IRS and FlyingThings3D for training surprisingly achieves the best or second-best accuracy among all the testing datasets.  
\begin{table}[htbp]
	\centering
	\caption{Benchmark results for disparity estimation on different synthetic datasets. Each cell shows the average EPE.}
	\label{tab:epe_synthetic}
	\begin{tabular}{|l||c|c|c|c|} \hline
		Method & IRS & FT3D  & Driving  & Sintel  \\ 
		& & \cite{mayer2016large} & \cite{mayer2016large} & \cite{butler2012naturalistic} \\ \hline\hline
		FADNet(FT3D) & 6.01 & \textbf{0.85} & 4.16 & 1.88 \\  
		FADNet(IRS) & 0.85 & 2.36 & 3.60  & 2.12 \\  
		FADNet(F+I) & \textbf{0.75} & 1.00  & \textbf{3.00} & \textbf{1.52} \\ \hline 
		GwcNet(FT3D) & 19.18 & \textbf{0.70} & 4.14 & 2.95 \\  
		GwcNet(IRS) & 4.29 & 1.99 &	4.00 & 1.98 \\  
		GwcNet(F+I) & \textbf{3.01} & 0.89 & \textbf{3.09} & \textbf{1.53}  \\ \hline 
	\end{tabular}
\end{table}

Then we explore the accuracy of FADNet/GwcNet trained by FlyingThings3D and IRS respectively on three real-world stereo datasets in Table \ref{tab:epe_real}. FADNet(IRS) outperforms FADNet(FT3D) on all of them, which indicates that the visual attributes of IRS are closer to the real-world captured. 
\begin{table}[ht]
	\caption{Benchmark results for disparity estimation on different natural datasets. Each cell shows the average EPE.}
	\centering
	\label{tab:epe_real}
	\begin{tabular}{|l||c|c|c|} \hline
		Method & Middlebury  & KITTI  & KITTI \\ 
		& (2014) \cite{scharstein2014high} 
		& (2012) \cite{kitti2012} 
		& (2015) \cite{kitti2015} 
		\\ \hline\hline
		FADNet(FT3D) & 2.68 & 1.41 & 1.74  \\ 
		FADNet(IRS) & 1.86 & \textbf{1.01} & 1.20 \\ 
		FADNet(F+I) & \textbf{1.47} & 1.04 & \textbf{1.15} \\ \hline
		GwcNet(FT3D) & 7.39 & 8.26 &  9.59 \\ 
		GwcNet(IRS) & 2.33 & 1.20 & 1.62 \\ 
		GwcNet(F+I) & \textbf{1.26} & \textbf{0.99} & \textbf{1.11}   \\ \hline
	\end{tabular}
	\vspace{-1.0 em}
\end{table}

\begin{figure}[htbp]
	\captionsetup[subfigure]{labelformat=empty, farskip=0pt}
	\centering
	\subfloat[]
	{
		\adjincludegraphics[width=0.22\linewidth,trim={{.2\width} 0 {.1\width} 0},clip]{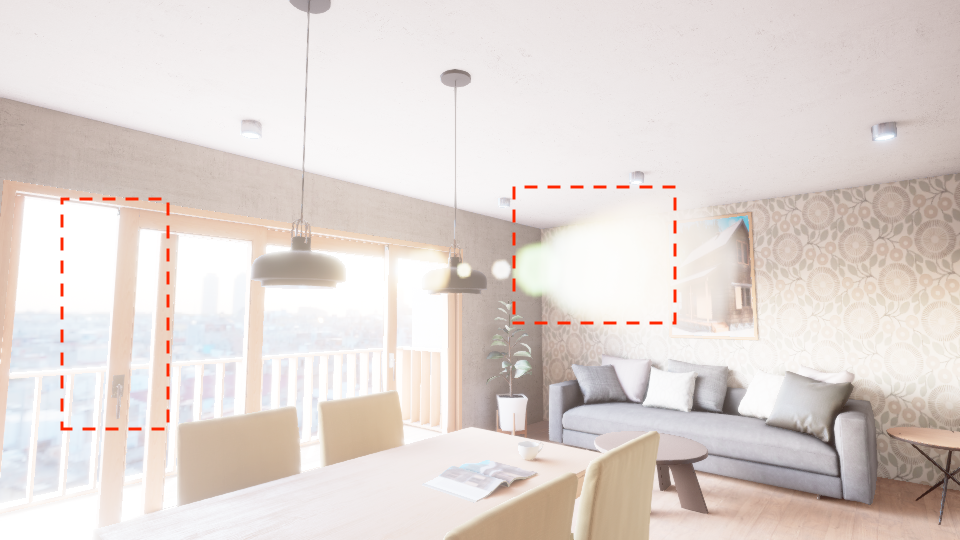}\label{fig:vt_bf_left_rgb}
	}
	\subfloat[]
	{
		\adjincludegraphics[width=0.22\linewidth,trim={{.2\width} 0 {.1\width} 0},clip]{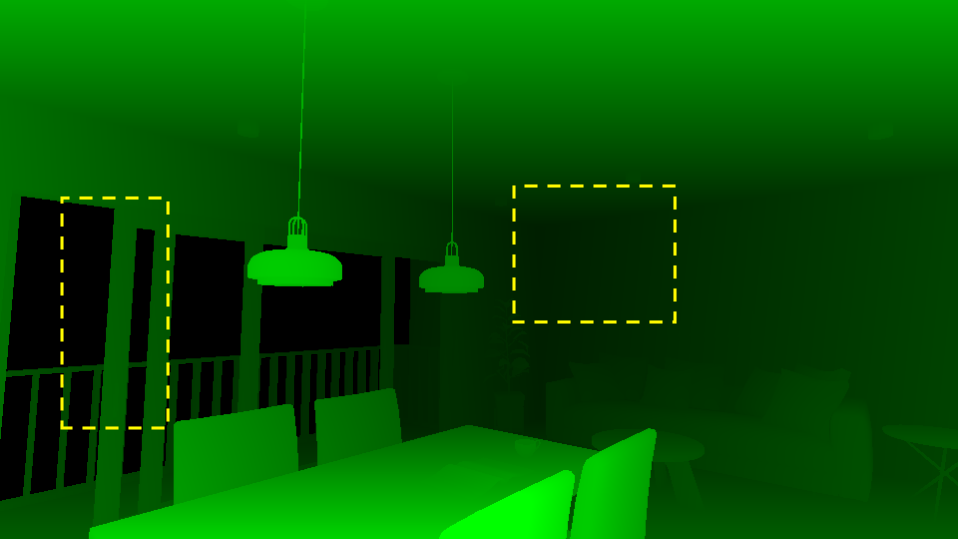}\label{fig:vt_bf_disp_gt}
	}
	\subfloat[]
	{
		\adjincludegraphics[width=0.22\linewidth,trim={{.2\width} 0 {.1\width} 0},clip]{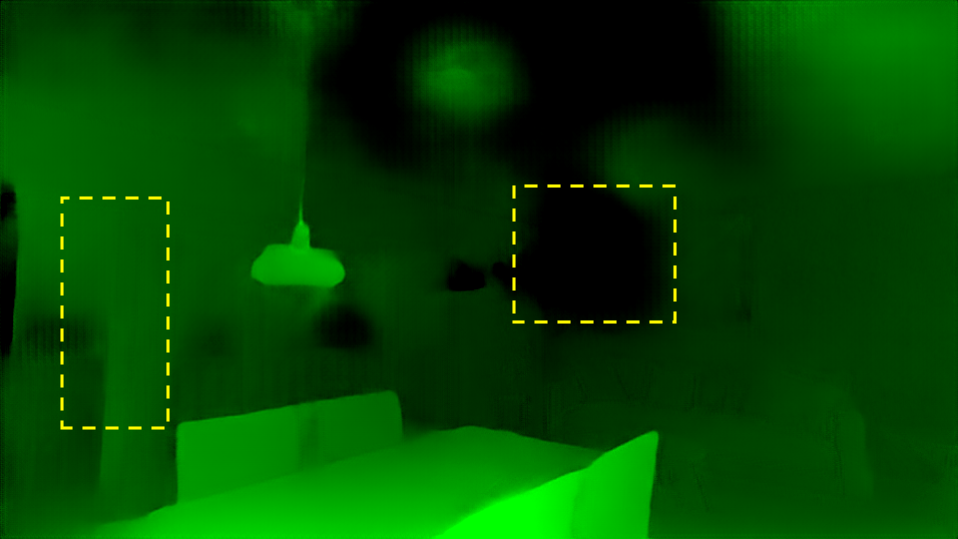}\label{fig:vt_bf_disp_irs}
	}
	\subfloat[]
	{
		\adjincludegraphics[width=0.22\linewidth,trim={{.2\width} 0 {.1\width} 0},clip]{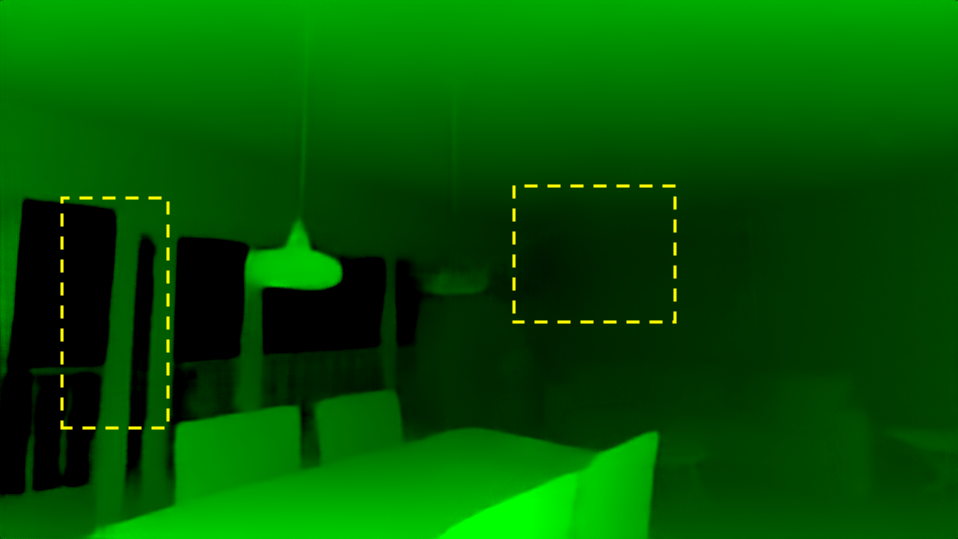}\label{fig:vt_bf_disp_sf}
	}
	\vspace{-1.0 em}
	\subfloat[Left RGB View]
	{
		\adjincludegraphics[width=0.22\linewidth,trim={{.2\width} 0 {.1\width} 0},clip]{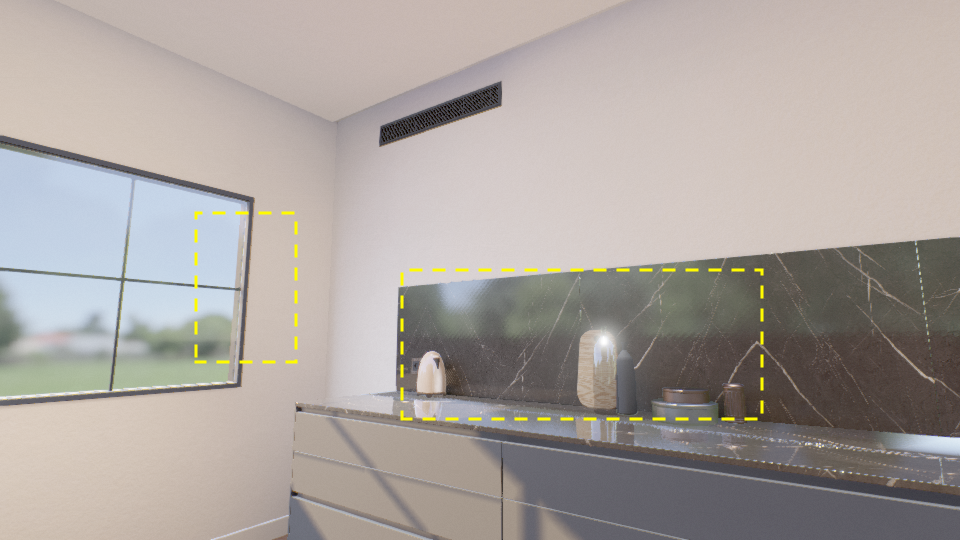}\label{fig:vt_rf_left_rgb}
	}
	\subfloat[Disparity GT]
	{
		\adjincludegraphics[width=0.22\linewidth,trim={{.2\width} 0 {.1\width} 0},clip]{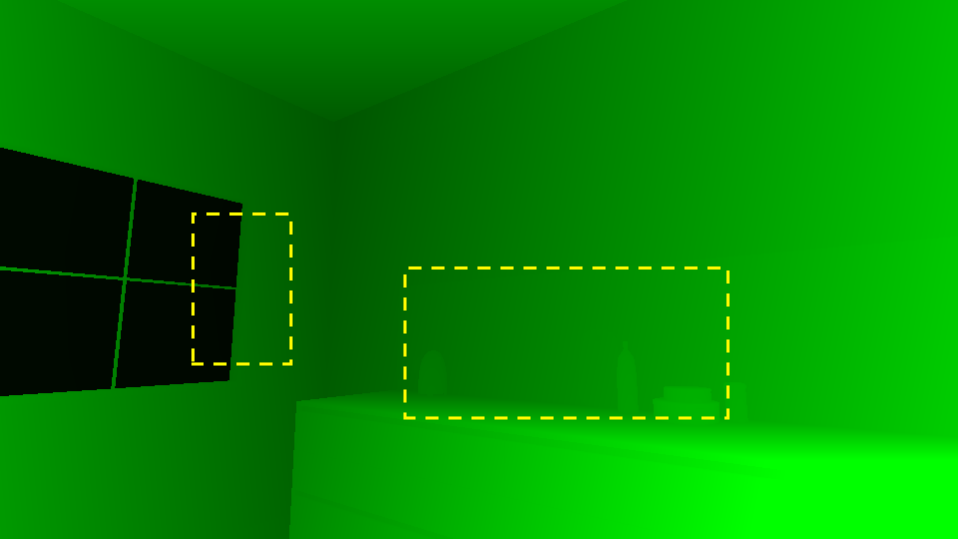}\label{fig:vt_rf_disp_gt}
	}
	\subfloat[FADNet(FD3T)]
	{
		\adjincludegraphics[width=0.22\linewidth,trim={{.2\width} 0 {.1\width} 0},clip]{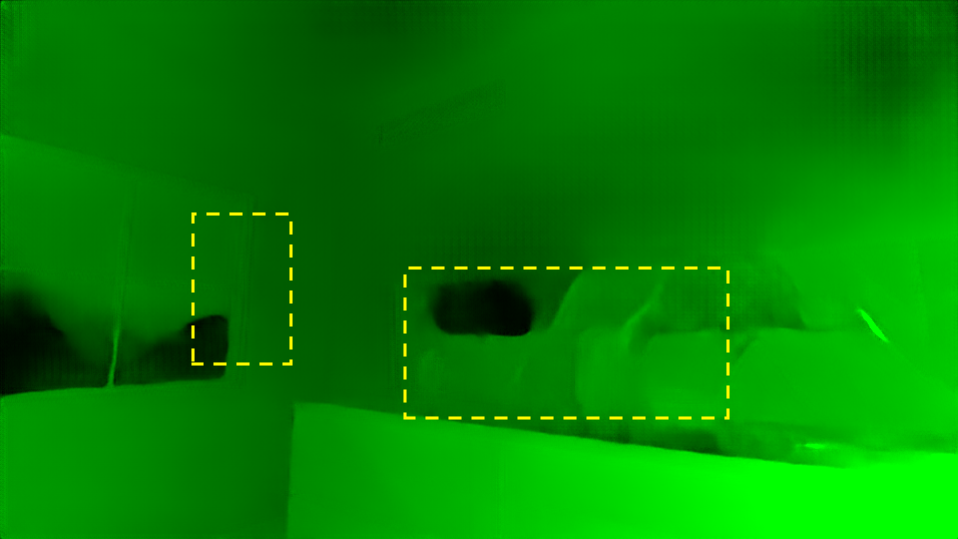}\label{fig:vt_rf_disp_irs}
	}
	\subfloat[FADNet(IRS)]
	{
		\adjincludegraphics[width=0.22\linewidth,trim={{.2\width} 0 {.1\width} 0},clip]{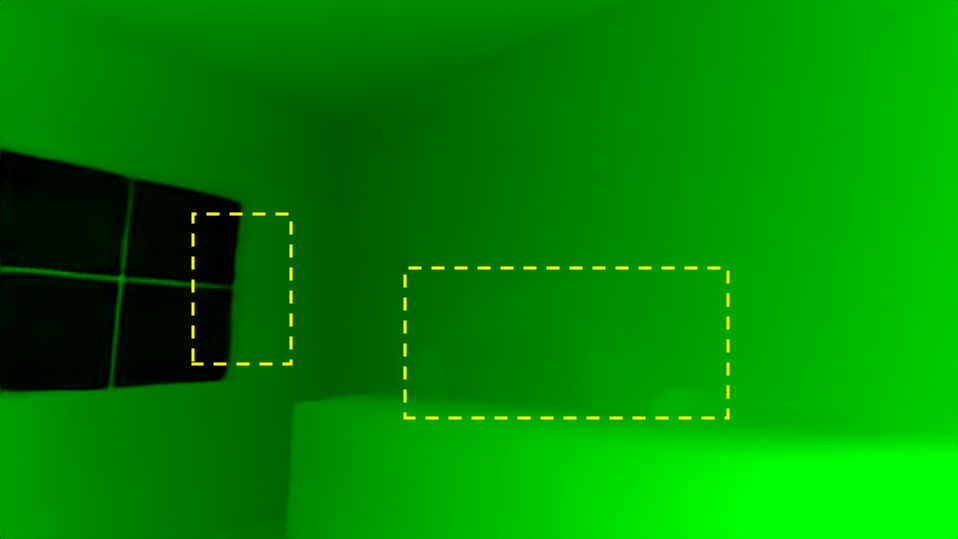}\label{fig:vt_rf_disp_sf}
	}
	\caption{Disparity Prediction Results on IRS Images.}
	\label{fig:disparity_on_synthetic}
\end{figure}

Fig. \ref{fig:disparity_on_synthetic} illustrates some examples of disparity maps predicted by FADNet(FT3D) and FADNet(IRS) on our IRS dataset. The first row shows a RGB image with lens flare and over-exposure. FADNet(FT3D) mistakes the lens flare patch as background and predicts very small values for those pixels, while FADNet(IRS) performs more robust results. 
The second row shows an image containing glass and mirrors. Since FlyingThings3D contains various of texture types, which tends to teach the network to learn disparity by feature matching, glass and mirrors can easily confuse FADNet(FT3D). However, FADNet trained on IRS can learn this kind of knowledge and produce much better results. 

\subsection{Benchmark Results for Surface Normal Estimation}
Table \ref{tab:normal_results} concludes the normal prediction accuracy of NormNetS, DFN-Net and DTN-Net. It reveals that DTN-Net achieves the best accuracy with a mean angle error of 10.64$^\circ$. Even over 74.1\% of pixels have low errors of no more than 11.25$^\circ$. Besides, two RGB-D based networks, DFN-Net and DTN-Net, considerably surpass the RGB based NormNetS, which implies that the disparity information can assist the network to predict better normal maps.  
\begin{figure}[ht]
	\captionsetup[subfigure]{labelformat=empty, farskip=0pt}
	\centering
	\subfloat[]
	{
		\adjincludegraphics[width=0.22\linewidth,trim={{.2\width} 0 {.1\width} 0},clip]{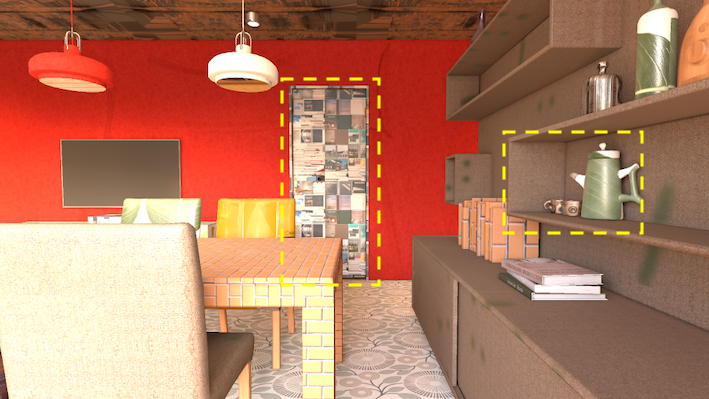}\label{fig:vt_1_left_rgb}
	}
	\subfloat[]
	{
		\adjincludegraphics[width=0.22\linewidth,trim={{.2\width} 0 {.1\width} 0},clip]{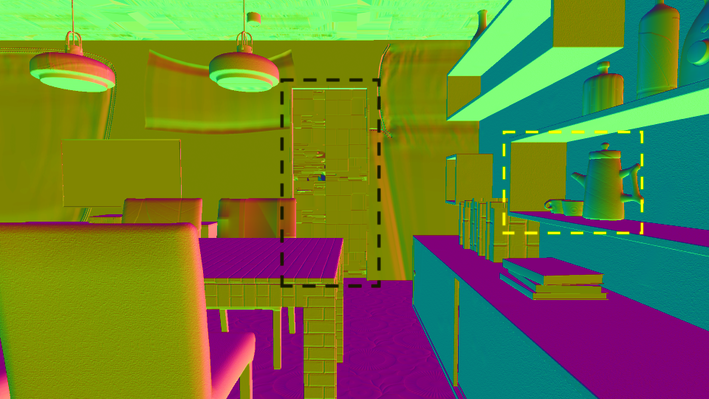}\label{fig:vt_1_left_normal_gt}
	}
	\subfloat[]
	{
		\adjincludegraphics[width=0.22\linewidth,trim={{.2\width} 0 {.1\width} 0},clip]{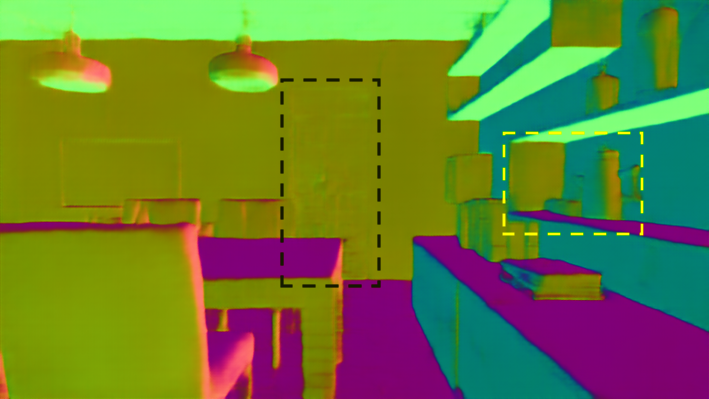}\label{fig:vt_ns_left_normal}
	}
	\subfloat[]
	{
		\adjincludegraphics[width=0.22\linewidth,trim={{.2\width} 0 {.1\width} 0},clip]{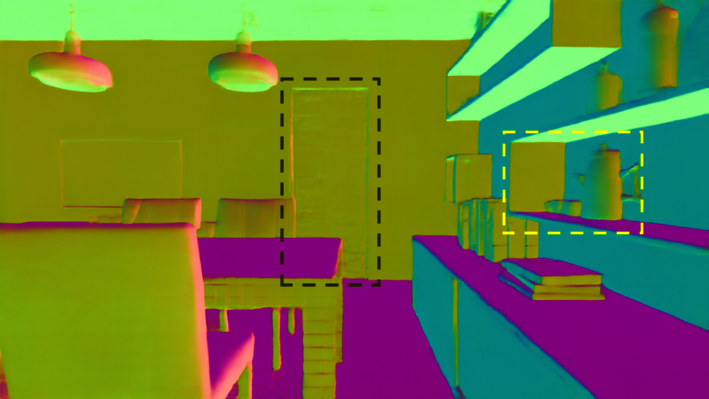}\label{fig:vt_dtn_left_normal}
	}
	\vspace{-1.0 em}
	\subfloat[Left RGB View]
	{
		\adjincludegraphics[width=0.22\linewidth,trim={{.2\width} 0 {.1\width} 0},clip]{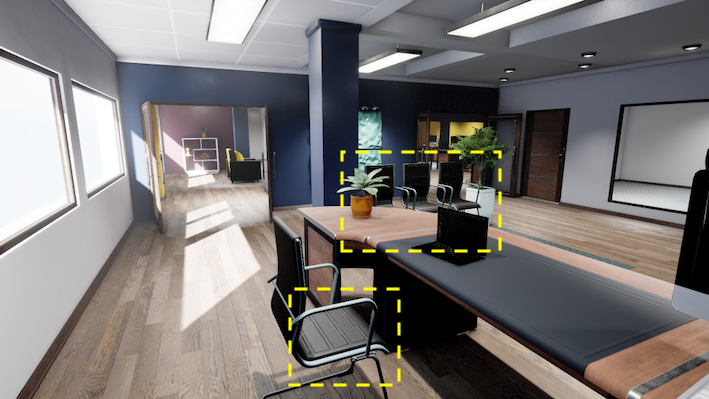}\label{fig:vt_3_left_rgb}
	}
	\subfloat[Normal GT]
	{
		\adjincludegraphics[width=0.22\linewidth,trim={{.2\width} 0 {.1\width} 0},clip]{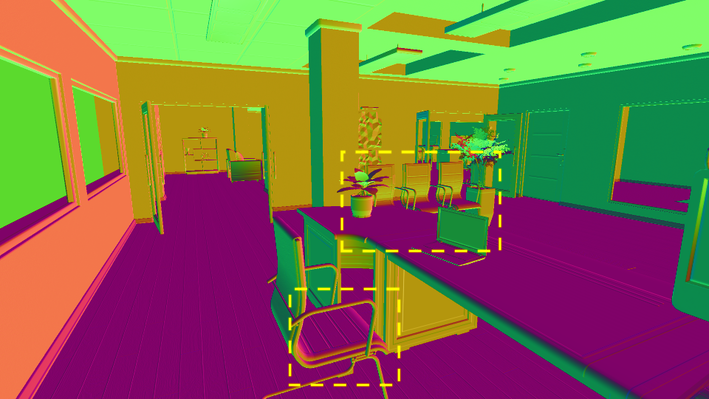}\label{fig:vt_3_left_normal_gt}
	}
	\subfloat[NormNetS]
	{
		\adjincludegraphics[width=0.22\linewidth,trim={{.2\width} 0 {.1\width} 0},clip]{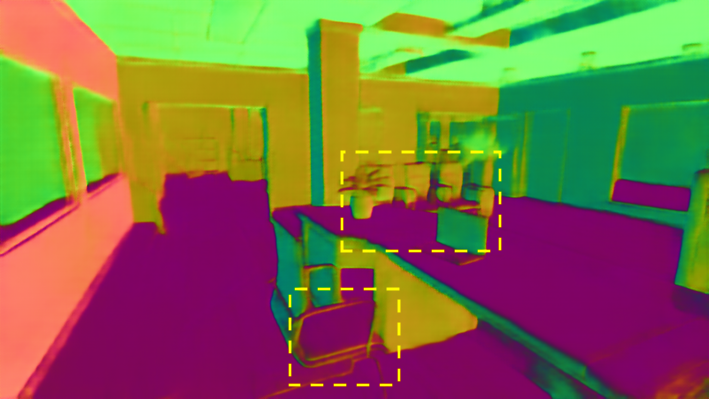}\label{fig:vt_3_ns_left_normal}
	}
	\subfloat[DTN-Net]
	{
		\adjincludegraphics[width=0.22\linewidth,trim={{.2\width} 0 {.1\width} 0},clip]{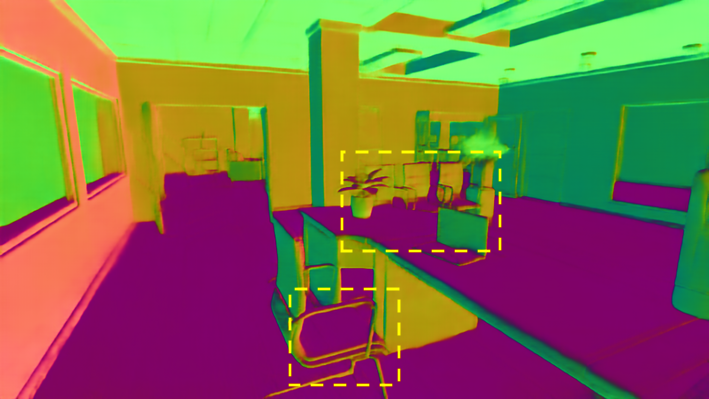}\label{fig:vt_3_dtn_left_normal}
	}
	\caption{Surface Normal Prediction Results on IRS Images.}
	\label{fig:normal_on_synthetic}
\end{figure}

Fig. \ref{fig:normal_on_synthetic} demonstrates some examples of normal maps predicted by NormNetS and DTN-Net. It is observed that DTN-Net provides better results in terms of object shape and smooth plane. However, some small objects in the image are still messed and cannot be recognized. We leave it as our future work of developing a better network training scheme. 
\begin{table}[ht]
	\centering
	\caption{Benchmark results for normal estimation of different networks on IRS.}
	\label{tab:normal_results}
	\begin{tabular}{|c|c|c|c|c|} \hline
		Method & mean  & $<$11.25$^\circ$ & $<$22.5$^\circ$ & $<$30$^\circ$ \\ \hline\hline
		NormNetS \cite{mayer2016large} & 13.93$^\circ$  & 65.7\% & 82.0\% & 87.0\% \\ \hline
		DFN-Net \cite{Zeng_2019_CVPR}& 12.81$^\circ$ & 69.1\% & 85.8\% & 89.8\% \\ \hline
		DTN-Net & \textbf{10.64$^\circ$} & \textbf{74.1\%} & \textbf{86.6\%} & \textbf{91.2\%} \\ \hline
	\end{tabular}
\end{table}

\subsection{Qualitative Results on Real World Captured Images}
\begin{figure}[htbp]
	\captionsetup[subfigure]{labelformat=empty, farskip=0pt}
	\centering
	\subfloat[]
	{
		\includegraphics[width=0.3\linewidth]{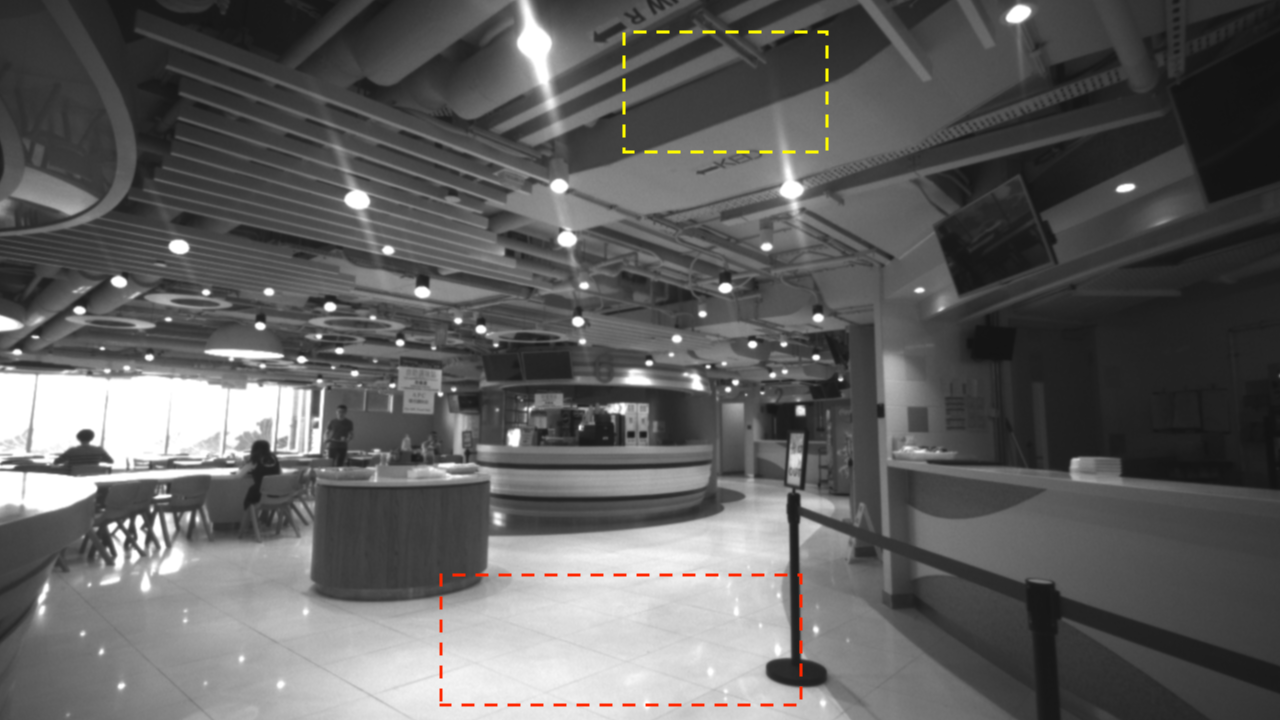}\label{fig:real_rf_left_rgb}
	}
	\subfloat[]
	{
		\includegraphics[width=0.3\linewidth]{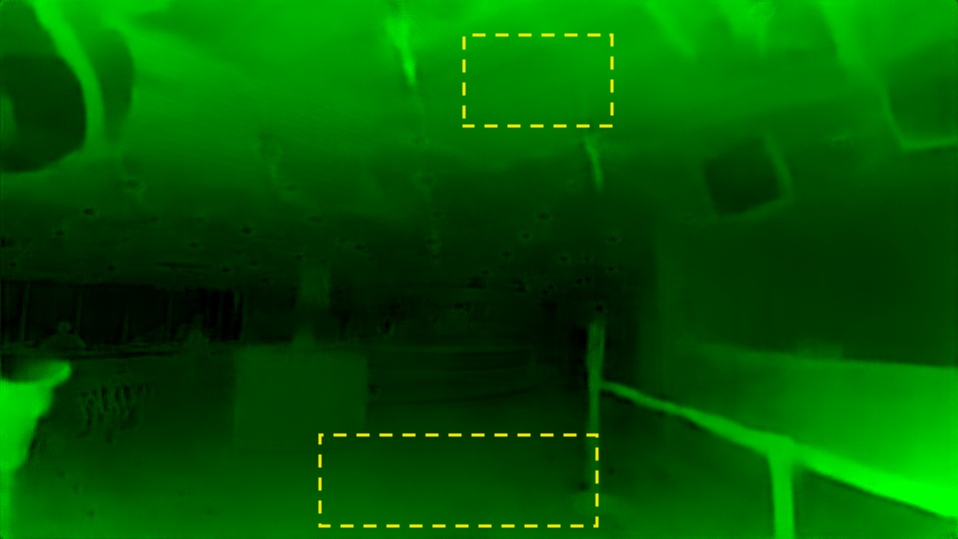}\label{fig:real_rf_disp_irs}
	}
	\subfloat[]
	{
		\includegraphics[width=0.3\linewidth]{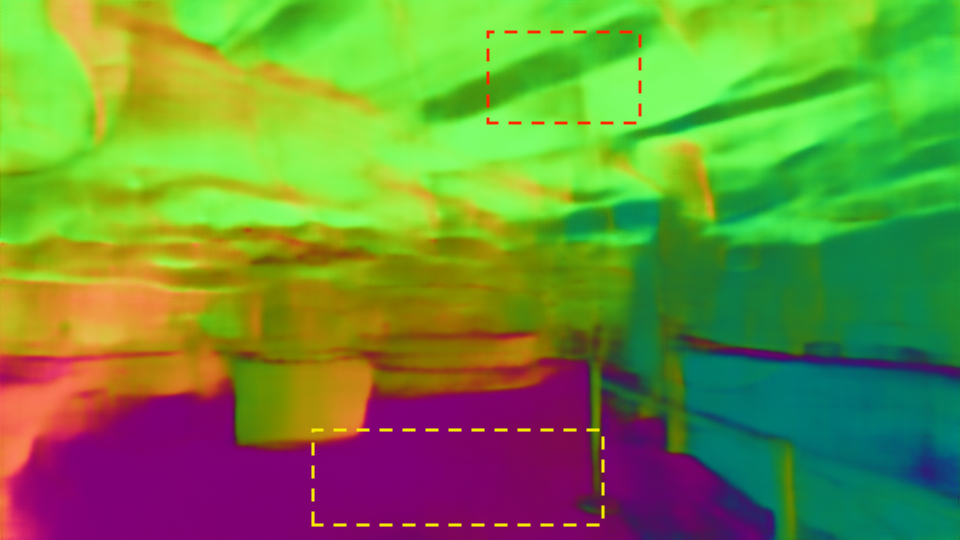}\label{fig:real_rf_normal_irs}
	}
	\vspace{-1.0 em}
	\subfloat[Left RGB View]
	{
		\includegraphics[width=0.3\linewidth]{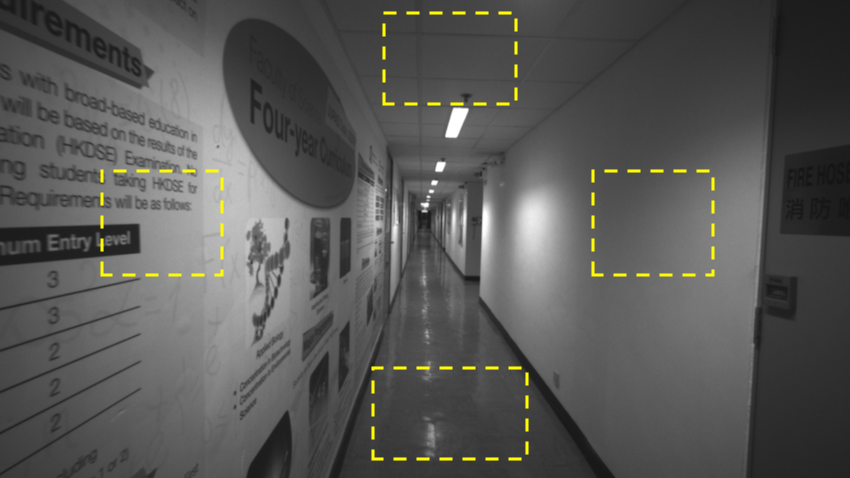}\label{fig:real_dis_left_rgb}
	}
	\subfloat[Disparity by FADNet]
	{
		\includegraphics[width=0.3\linewidth]{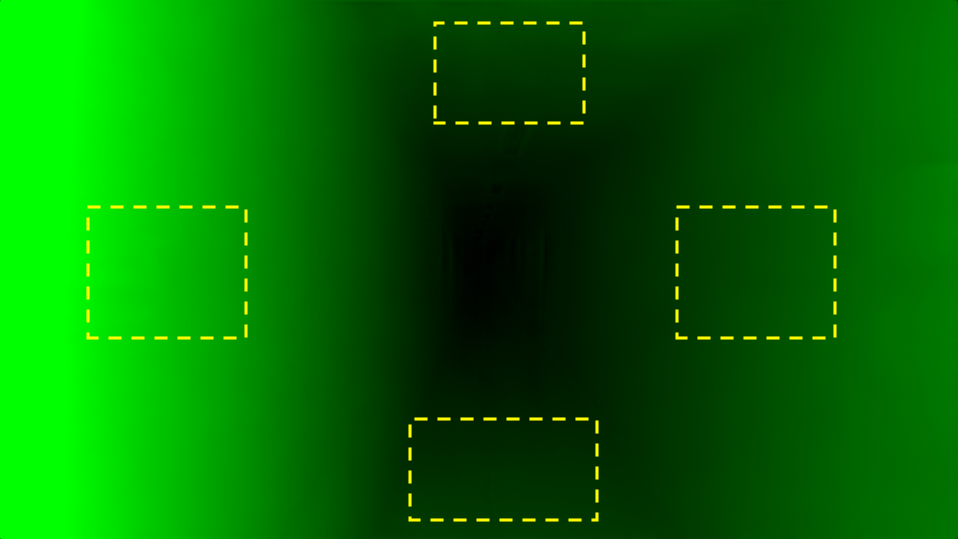}\label{fig:real_dis_disp_irs}
	}
	\subfloat[Normal by DTN-Net]
	{
		\includegraphics[width=0.3\linewidth]{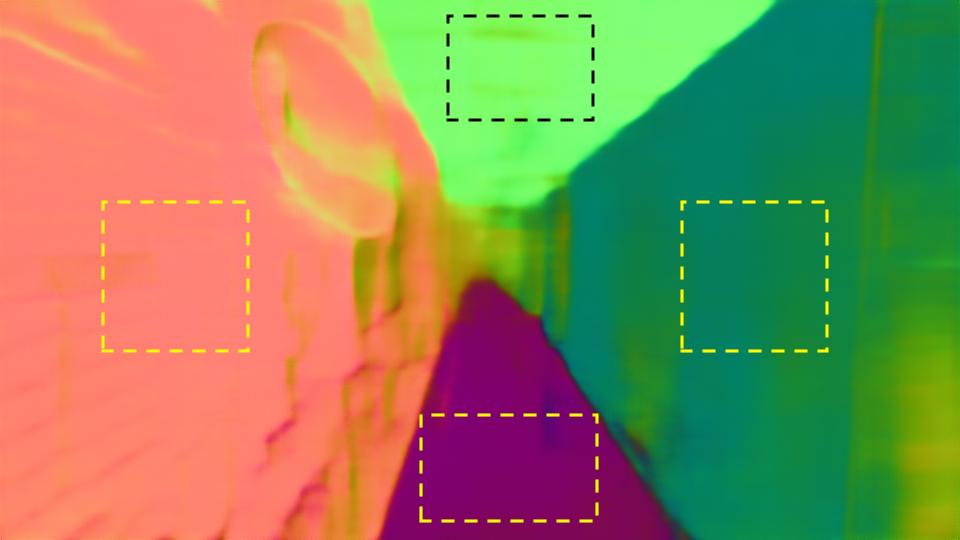}\label{fig:real_dis_normal_irs}
	}
	\caption{Disparity and Surface Normal Prediction Results of FADNet and DTN-Net on Real World Images.}
	\label{fig:disparity_on_real}
\end{figure}

We evaluate the performance of FADNet and DTN-Net trained by IRS on the real world data. We use Intel RealSense D435i stereo camera to capture indoor images of offices and canteens and then apply those two models to predict their disparity and normal maps. Fig. \ref{fig:disparity_on_real} illustrates two examples. 
The first example contains the mirror-surface ground which reflects the light. It is surprising that FADNet and DTN-Net assign those regions with a smooth disparity and normal map, which is reasonable for a plane. The second example shows a wall with rich texture and a long aisle. FADNet and DTN-Net can still notice the flatness of walls, floors and ceiling, and produce satisfying disparity and normal maps. 

\section{Conclusion}\label{sec:conclusion}
In this paper, we propose IRS, a large-scale synthetic stereo dataset for indoor robotics targeted at disparity and surface normal estimation. IRS covers a wide range of indoor scenes, including home, office, restaurant and store, and are remarkably close to the real world captured images in the aspects of brightness changes, light reflection/transmission, lens flare, etc. To illustrate the usage and functionality of IRS, we first compare the data distribution of IRS to other stereo datasets to analyze the classical visual attributes of indoor scenes. We then develop two deep models, RD-Net and DTN-Net, to respectively predict the disparity and surface normal maps from stereo images. On the one hand, RD-Net trained with IRS produces the best accuracy for common indoor scenes and existing real-world stereo datasets, and performs decent generalization for both synthetic and natural stereo data. On the other hand, by refining the initial normal map transformed from the predicted disparity, DTN-Net predicts better final normal maps than the existing RGB-based and depth-fusion models.

\bibliographystyle{IEEEtran}
\bibliography{main.bbl}

\clearpage
\section*{Appendix}

\subsection{Model Design}
Our proposed DTN-Net is composed of two sub-networks, RD-Net \cite{wang2020fadnet} and NormNetS. RD-Net takes the stereo RGB images as input to predict the left disparity map. We then transform the predicted disparity map into the initial surface normal map and feed it as part of the input of NormNetS. 
\subsubsection{RD-Net for Disparity Estimation}
The parameters of RD-Net are listed in Table \ref{tab:rdnet}. RD-Net is composed of four parts. The first shared module is the layers used to extract and shrink the feature maps of the left and right images respectively. The weights of this module are shared between the input left and right images. Then the second left-right correlation module connects the final left and right feature maps from the shared module, and calculates their correspondence. The third encoder module further extracts the high-level features from the correspondence maps, and predicts the disparity map of the smallest scale, $\frac{1}{64}$H $\times$ $\frac{1}{64}$W. The final decoder module reversely reconstructs the feature maps, and predicts the disparity maps of different scales, from $\frac{1}{32}$H $\times$ $\frac{1}{32}$W to H $\times$ W. Besides, some feature maps in the encoder module are also connected to those in the decoder module, which enhances the details of predicted disparity with the help of shallow features. Different from the six convolution layers in the encoder module of DispFulNet \cite{crl}, we substitude them with three dual residual blocks, which are introduced in \cite{wang2020fadnet}. 
\begin{table}[htbp]
	\centering
	\begin{threeparttable}
		\caption{Parameters of RD-Net used for disparity estimation.}
		\label{tab:rdnet}
		\scriptsize{
			\begin{tabular}
				{|c|c|c|c|} \hline
				Name & Layer Setting & Output Dimension & Input \\ \hline\hline
				img(l/r) & - & H $\times$ W $\times$ 3 & - \\ \hline\hline
				\multicolumn{4}{|c|}{Shared Module} \\ \hline
				conv1 & 7 $\times$ 7 $\times$ 64 & $\frac{1}{2}$H $\times$ $\frac{1}{2}$W $\times$ 64 & img(l/r) \\ \hline
				conv2 & 3 $\times$ 3 $\times$ 128 & $\frac{1}{4}$H $\times$ $\frac{1}{4}$W $\times$ 128 & conv1 \\ \hline
				conv3 & 3 $\times$ 3 $\times$ 256 & $\frac{1}{8}$H $\times$ $\frac{1}{8}$W $\times$ 256 & conv2 \\ \hline\hline
				\multicolumn{4}{|c|}{Left-Right Correlation Module} \\ \hline
				corr & D=40 & $\frac{1}{8}$H $\times$ $\frac{1}{8}$W $\times$ 40 & conv3(l), conv3(r)\\ \hline
				conv3\_redir & 3 $\times$ 3 $\times$ 32 & $\frac{1}{8}$H $\times$ $\frac{1}{8}$W $\times$ 32 & conv3(l) \\ \hline
				concat\_corr & - & $\frac{1}{8}$H $\times$ $\frac{1}{8}$W $\times$ 72 & corr, conv3\_redir \\ \hline
				conv3\_f & 3 $\times$ 3 $\times$ 256 & $\frac{1}{8}$H $\times$ $\frac{1}{8}$W $\times$ 256 & concat\_corr \\ \hline\hline
				\multicolumn{4}{|c|}{Encoder Module} \\ \hline
				drb4 & channel=512 & $\frac{1}{16}$H $\times$ $\frac{1}{16}$W $\times$ 512 & conv3\_f \\ \hline
				drb5 & channel=512 & $\frac{1}{32}$H $\times$ $\frac{1}{32}$W $\times$ 512 & drb4 \\ \hline
				drb6 & channel=1024 & $\frac{1}{64}$H $\times$ $\frac{1}{64}$W $\times$ 1024 & drb5 \\ \hline
				pr6 & 3 $\times$ 3 $\times$ 1 & $\frac{1}{64}$H $\times$ $\frac{1}{64}$W $\times$ 1 & drb6 \\ \hline\hline
				\multicolumn{4}{|c|}{Decoder Module} \\ \hline
				upflow6 & 4 $\times$ 4 $\times$ 1 & $\frac{1}{32}$H $\times$ $\frac{1}{32}$W $\times$ 1 & pr6 \\ \hline
				upconv5 & 4 $\times$ 4 $\times$ 512 & $\frac{1}{32}$H $\times$ $\frac{1}{32}$W $\times$ 512 & drb6 \\ \hline
				concat5 & - & $\frac{1}{32}$H $\times$ $\frac{1}{32}$W $\times$ 1025 & upflow6, upconv5, drb5 \\ \hline
				iconv5 & 3 $\times$ 3 $\times$ 512 & $\frac{1}{32}$W $\times$ 512 & concat5 \\ \hline
				pr5 & 3 $\times$ 3 $\times$ 1 & $\frac{1}{32}$H $\times$ $\frac{1}{32}$W $\times$ 1 & iconv5 \\ \hline
				upflow5 & 4 $\times$ 4 $\times$ 1 & $\frac{1}{16}$H $\times$ $\frac{1}{16}$W $\times$ 1 & pr5 \\ \hline
				upconv4 & 4 $\times$ 4 $\times$ 256 & $\frac{1}{16}$H $\times$ $\frac{1}{16}$W $\times$ 256 & iconv5 \\ \hline
				concat4 & - & $\frac{1}{16}$H $\times$ $\frac{1}{16}$W $\times$ 769 & upflow5, upconv4, drb4 \\ \hline
				iconv4 & 3 $\times$ 3 $\times$ 256 & $\frac{1}{16}$H $\times$ $\frac{1}{16}$W $\times$ 256 & concat4 \\ \hline
				pr4 & 3 $\times$ 3 $\times$ 1 & $\frac{1}{16}$H $\times$ $\frac{1}{16}$W $\times$ 1 & iconv4 \\ \hline
				upflow4 & 4 $\times$ 4 $\times$ 1 & $\frac{1}{8}$H $\times$ $\frac{1}{8}$W $\times$ 1 & pr4 \\ \hline
				upconv3 & 4 $\times$ 4 $\times$ 128 & $\frac{1}{8}$H $\times$ $\frac{1}{8}$W $\times$ 128 & iconv4 \\ \hline
				concat3 & - & $\frac{1}{8}$H $\times$ $\frac{1}{8}$W $\times$ 385 & upflow4, upconv3, conv3\_f \\ \hline
				iconv3 & 3 $\times$ 3 $\times$ 128 & $\frac{1}{8}$H $\times$ $\frac{1}{8}$W $\times$ 128 & concat3 \\ \hline
				pr3 & 3 $\times$ 3 $\times$ 1 & $\frac{1}{8}$H $\times$ $\frac{1}{8}$W $\times$ 1 & iconv3 \\ \hline
				upflow3 & 4 $\times$ 4 $\times$ 1 & $\frac{1}{4}$H $\times$ $\frac{1}{4}$W $\times$ 1 & pr3 \\ \hline
				upconv2 & 4 $\times$ 4 $\times$ 64 & $\frac{1}{4}$H $\times$ $\frac{1}{4}$W $\times$ 64 & iconv3 \\ \hline
				concat2 & - & $\frac{1}{4}$H $\times$ $\frac{1}{4}$W $\times$ 193 & upflow3, upconv2, conv2(l) \\ \hline
				iconv2 & 3 $\times$ 3 $\times$ 64 & $\frac{1}{4}$H $\times$ $\frac{1}{4}$W $\times$ 64 & concat2 \\ \hline
				pr2 & 3 $\times$ 3 $\times$ 1 & $\frac{1}{4}$H $\times$ $\frac{1}{4}$W $\times$ 1 & iconv2 \\ \hline
				upflow2 & 4 $\times$ 4 $\times$ 1 & $\frac{1}{2}$H $\times$ $\frac{1}{2}$W $\times$ 1 & pr2 \\ \hline
				upconv1 & 4 $\times$ 4 $\times$ 32 & $\frac{1}{2}$H $\times$ $\frac{1}{2}$W $\times$ 32 & iconv2 \\ \hline
				concat1 & - & $\frac{1}{2}$H $\times$ $\frac{1}{2}$W $\times$ 97 & upflow2, upconv1, conv1(l) \\ \hline
				iconv1 & 3 $\times$ 3 $\times$ 32 & $\frac{1}{2}$H $\times$ $\frac{1}{2}$W $\times$ 32 & concat1 \\ \hline
				pr1 & 3 $\times$ 3 $\times$ 1 & $\frac{1}{2}$H $\times$ $\frac{1}{2}$W $\times$ 1 & iconv1 \\ \hline
				upflow1 & 4 $\times$ 4 $\times$ 1 & H $\times$ W $\times$ 1 & pr1 \\ \hline
				upconv0 & 4 $\times$ 4 $\times$ 16 & H $\times$ W $\times$ 16 & iconv1 \\ \hline
				concat0 & - & H $\times$ W $\times$ 20 & upflow1, upconv0, img(l)\\ \hline
				iconv0 & 3 $\times$ 3 $\times$ 16 & H $\times$ W $\times$ 16 & concat0 \\ \hline	
				pr0 & 3 $\times$ 3 $\times$ 1 & H $\times$ W $\times$ 1 & iconv0 \\ \hline
			\end{tabular}
		}	
		\begin{tablenotes}
			\item[]Note:``conv" and ``iconv" indicate the standard convolution layer. ``corr" indicates the correlation layer. ``drb" indicates the dual residual block. ``pr" indicates the convolution layer used for predicting the disparity. ``concat" indicates the concatenate operation. ``upflow" and ``upconv" indicate the transposed convolution layer. ``l" and ``r" indicate the images and feature maps of the left and right view. 
		\end{tablenotes}
	\end{threeparttable}
\end{table}
\subsubsection{DTN-Net for Surface Normal Estimation}
DTN-Net is composed of RD-Net and NormNetS. The parameters of NormNetS are listed in Table \ref{tab:normnets}. Similar to DispNet \cite{mayer2016large}, NormNetS is composed of two parts, the encoder module and the decoder module. We also replace the original convolution layers in the encoder part by the dual residual blocks. Besides, since the normal map is three-channel, $(x,y,z)$ for the 3D coordinates, we post-process the output normal map by normalizing it along channels for each pixel, which leads to a unit vector. 
To construct DTN-Net, we connect RD-Net and NormNetS as follows. RD-Net firstly produces the final disparity map of the original size. Then this disparity map is converted into the initial coarse normal map. NormNetS finally concatenates the left and right images, the initial normal map and the final disparity map as input, and predicts the final delicate normal map.

\begin{table}[ht]
	\centering
	\begin{threeparttable}
		\caption{Parameters of NormNetS used for surface normal estimation.}
		\label{tab:normnets}
		\scriptsize{
			\begin{tabular}	
				{|c|c|c|c|} \hline
				Name & Layer Setting & Output Dimension & Input \\ \hline\hline
				img(l) + img(r) & - & H $\times$ W $\times$ 6 & - \\ \hline\hline
				\multicolumn{4}{|c|}{Encoder Module} \\ \hline
				conv1 & 7 $\times$ 7 $\times$ 64 & $\frac{1}{2}$H $\times$ $\frac{1}{2}$W $\times$ 64 & img(l) + img(r) \\ \hline
				conv2 & 3 $\times$ 3 $\times$ 128 & $\frac{1}{4}$H $\times$ $\frac{1}{4}$W $\times$ 128 & conv1 \\ \hline
				conv3 & 3 $\times$ 3 $\times$ 256 & $\frac{1}{8}$H $\times$ $\frac{1}{8}$W $\times$ 256 & conv2 \\ \hline
				conv3\_1 & 3 $\times$ 3 $\times$ 256 & $\frac{1}{8}$H $\times$ $\frac{1}{8}$W $\times$ 256 & conv3 \\ \hline		
				drb4 & channel=512 & $\frac{1}{16}$H $\times$ $\frac{1}{16}$W $\times$ 512 & conv3\_f \\ \hline
				drb5 & channel=512 & $\frac{1}{32}$H $\times$ $\frac{1}{32}$W $\times$ 512 & drb4 \\ \hline
				drb6 & channel=1024 & $\frac{1}{64}$H $\times$ $\frac{1}{64}$W $\times$ 1024 & drb5 \\ \hline
				\multicolumn{4}{|c|}{Decoder Module} \\ \hline
				upconv5 & 4 $\times$ 4 $\times$ 512 & $\frac{1}{32}$H $\times$ $\frac{1}{32}$W $\times$ 512 & drb6 \\ \hline
				concat5 & - & $\frac{1}{32}$H $\times$ $\frac{1}{32}$W $\times$ 1024 & upconv5, drb5 \\ \hline
				iconv5 & 3 $\times$ 3 $\times$ 512 & $\frac{1}{32}$W $\times$ 512 & concat5 \\ \hline
				upconv4 & 4 $\times$ 4 $\times$ 256 & $\frac{1}{16}$H $\times$ $\frac{1}{16}$W $\times$ 256 & iconv5 \\ \hline
				concat4 & - & $\frac{1}{16}$H $\times$ $\frac{1}{16}$W $\times$ 768 & upconv4, drb4 \\ \hline
				iconv4 & 3 $\times$ 3 $\times$ 256 & $\frac{1}{16}$H $\times$ $\frac{1}{16}$W $\times$ 256 & concat4 \\ \hline
				upconv3 & 4 $\times$ 4 $\times$ 128 & $\frac{1}{8}$H $\times$ $\frac{1}{8}$W $\times$ 128 & iconv4 \\ \hline
				concat3 & - & $\frac{1}{8}$H $\times$ $\frac{1}{8}$W $\times$ 384 & upconv3, conv3\_f \\ \hline
				iconv3 & 3 $\times$ 3 $\times$ 128 & $\frac{1}{8}$H $\times$ $\frac{1}{8}$W $\times$ 128 & concat3 \\ \hline
				upconv2 & 4 $\times$ 4 $\times$ 64 & $\frac{1}{4}$H $\times$ $\frac{1}{4}$W $\times$ 64 & iconv3 \\ \hline
				concat2 & - & $\frac{1}{4}$H $\times$ $\frac{1}{4}$W $\times$ 192 & upconv2, conv2 \\ \hline
				iconv2 & 3 $\times$ 3 $\times$ 64 & $\frac{1}{4}$H $\times$ $\frac{1}{4}$W $\times$ 64 & concat2 \\ \hline
				upconv1 & 4 $\times$ 4 $\times$ 32 & $\frac{1}{2}$H $\times$ $\frac{1}{2}$W $\times$ 32 & iconv2 \\ \hline
				concat1 & - & $\frac{1}{2}$H $\times$ $\frac{1}{2}$W $\times$ 96 & upconv1, conv1 \\ \hline
				iconv1 & 3 $\times$ 3 $\times$ 32 & $\frac{1}{2}$H $\times$ $\frac{1}{2}$W $\times$ 32 & concat1 \\ \hline
				upconv0 & 4 $\times$ 4 $\times$ 16 & H $\times$ W $\times$ 16 & iconv1 \\ \hline
				concat0 & - & H $\times$ W $\times$ 19 & upconv0, img(l)\\ \hline
				iconv0 & 3 $\times$ 3 $\times$ 16 & H $\times$ W $\times$ 16 & concat0 \\ \hline	
				pn & 3 $\times$ 3 $\times$ 3 & H $\times$ W $\times$ 1 & iconv0 \\ \hline
			\end{tabular}
		}	
		\begin{tablenotes}
			\item[]Note:``conv" and ``iconv" indicate the standard convolution layer. ``drb" indicates the dual residual block. ``pn" indicates the convolution layer used for predicting the surface normal. ``concat" indicates the concatenate operation. ``upconv" indicates the transposed convolution layer. ``l" and ``r" indicate the left and right images. We post-process the output 3-channel normal map by normalizing it along channels for each pixel.
		\end{tablenotes}
	\end{threeparttable}
	\vspace{-1.2 em}
\end{table}

\subsubsection{Disparity-to-Normal Module (D2N)}
Surface normal is determined by local surface tangent plane of neighboring 3D points, which can be roughly estimated from depth information. We adopt this rule to the transformation between disparity and normal. 
We adopt the classical pinhole camera model in our geometric system. We denote $(u_{i}, v_{i})$ at the location of pixel $i$ in the 2D image. Its corresponding location in 3D space is $(x_{i}, y_{i}, z_{i})$, where $z_{i}$ is the depth. Based on the geometry of perspective projection, we obtain 
\begin{align}
x_{i}=(u_{i} - c_{x}) * z_{i} / f_{x} \nonumber \\ 
y_{i}=(v_{i} - c_{y}) * z_{i} / f_{y} \label{eq:3d_pos}
\end{align}
where $f_{x}$ and $f_{y}$ are the focal length along the $x$ and $y$ directions respectively. $c_{x}$ and $c_{y}$ are coordinates of the principal points. In a 2D image, if we assume that the surrounding pixel $j$ are in the same plane of the pixel $i$, we observe that the relationship between their 3D space locations and surface normal vectors can be described by
\begin{align}
(z_j-z_i)/(x_i-x_j)=n_{ix}/n_{iz} \nonumber \\
(z_j-z_i)/(y_i-y_j)=n_{iy}/n_{iz} \label{eq:normal}
\end{align}
where $(n_{ix}, n_{iy}, n_{iz})$ is the surface normal vector of pixel $i$.
Given a pair of rectified images, the disparity of pixel $i$ on the left image is the offset $d_i$ of its location at $(u_i-d_i, v_i)$ on the right image. The depth $z_i$ can be computed as,
\begin{align}
z_i=f_x * b / d_i \label{eq:depth}
\end{align}
where $b$ is the baseline of stereo camera. \\
Combining Eq. \eqref{eq:3d_pos}\eqref{eq:depth} we can figure out the 3D position $(x_{i}, y_{i}, z_{i})$ of pixel $i$. After that, for any pixel $j\in \mathbf{M}_i$, which means pixels surround the pixel $i$, we can gain an estimate normal on pixel $i$ through Eq. \ref{eq:normal} as 
\begin{align}
\mathbf{n}_{i}&=({n}_{ix},{n}_{iy},{n}_{iz}) \nonumber \\
&=(\frac{z_j-z_i}{x_j-x_i}, \frac{z_j-z_i}{y_j-y_i}, -1) 
\end{align}
To ease model training, we simply average the estimated surface normal vectors of all $j$ in our implementation to obtain the final normal vector of pixel $i$, which is calculated by Eq. \eqref{eq:mean_normal}. 
\begin{align}
\hat{\mathbf{n}}_{i}=\mathbf{MEAN}_{j\in \mathbf{M}_i}(\mathbf{n}_{ji}) \label{eq:mean_normal}
\end{align}
where $\mathbf{n}_{ji}$ is an estimate from pixel $j$. 

\subsection{Loss Function Design}\label{subsec:loss}
Given a pair of stereo RGB images, our RD-Net takes them as input and produces seven disparity maps at different scales. Assume that the input image size is $H \times W$. The dimension of the seven scales of the output disparity maps are $H \times W$, $\frac{1}{2}H \times \frac{1}{2}W$, $\frac{1}{4}H \times \frac{1}{4}W$, $\frac{1}{8}H \times \frac{1}{8}W$, $\frac{1}{16}H \times \frac{1}{16}W$, $\frac{1}{32}H \times \frac{1}{32}W$, and $\frac{1}{64}H \times \frac{1}{64}W$ respectively. To train RD-Net in an end-to-end manner, we adopt the pixel-wise smooth L1 loss between the predicted disparity map and the ground truth using  
\begin{align}
L_s(d_s, \hat{d_s})=\frac{1}{N}\sum_{i=1}^{N}{smooth}_{L_1}(d_{s}^i - \hat{d_{s}^i}) \label{eq:smooth_l1}
\end{align}
where $N$ is the number of pixels of the disparity map, $d_s^i$ is the $i^th$ element of $d_s\in \mathcal{R}^N$ and 
\begin{align}
{smooth}_{L_1}(x)=
\begin{cases}
0.5x^2,& \text{if } |x| < 1\\
|x|-0.5,              & \text{otherwise}.
\end{cases}
\end{align}

Note that $d_s$ is the ground truth disparity of scale $\frac{1}{2^s}$ and $\hat{d_s}$ is the predicted disparity of scale $\frac{1}{2^s}$. The loss function is separately applied in the seven scales of outputs, which generates seven loss values. The loss values are then accumulated with loss weights. 
The loss weight scheduling technique which is initially proposed in \cite{mayer2016large} is useful to learn the disparity in a coarse-to-fine manner. Instead of just switching on/off the losses of different scales, we apply different non-zero weight groups for tackling different scale of disparity. Let $w_s$ denote the weight for the loss of the scale of $s$. The final disparity loss function $L_d$ is
\begin{equation}
L_d=\sum_{s=0}^{6}w_sL_s(d_s,\hat{d_s})
\end{equation}

There are seven scales of disparity maps. At the beginning, we assign low-value weights for those large scale disparity maps to learn the coarse features. Then we increase the loss weights of large scales to let the network gradually learn the finer features. Finally, we deactivate all the losses except the final predict one of the original input size. 
The surface normal loss $L_n$ is
\begin{equation}
L_n=\frac{1}{N}\sum_{i=1}^{N}||\mathbf{n}_i - \mathbf{n}_{i}^{gt}||_{2}^{2} \label{eq:l_normal}
\end{equation}
where $\mathbf{n}_i$ and $\mathbf{n}_{i}^{gt}$ are respectively the predicted surface normal and the ground truth. Thus, the overall loss function is $L = L_d + L_n$. 

\subsection{Training Settings}
We implemented FADNet, GwcNet, NormNetS, DFN-Net and DTN-Net using PyTorch. All the models were end-to-end trained with Adam (${\beta}_1=0.9, {\beta}_2=0.999$). We performed color normalization with the mean and variation of ImageNet \cite{krizhevsky2012imagenet} for data preprocessing. During training, images were randomly cropped to size $H = 384$ and $W = 768$. The batch size was set to 16 on four Nvidia Titan X Pascal GPUs. For FADNet and DTN-Net, we totally train the models for 90 epoches and follow the multi-scale loss weight scheme in \cite{wang2020fadnet}.  The learning rate was initialized as $10^{-4}$ at the beginning of each round and decreased by half every 10 epoches. For NormNetS, we train it for 30 epoches.  The learning rate was initialized as $10^{-4}$ and decreased by half every 10 epoches.

\subsection{Visual Examples}\label{sec:visual}
\begin{figure}[htbp]
	\captionsetup[subfigure]{labelformat=empty, farskip=0pt}
	\centering
	\subfloat[]
	{
		\adjincludegraphics[width=0.175\linewidth,trim={{.2\width} 0 {.1\width} 0},clip]{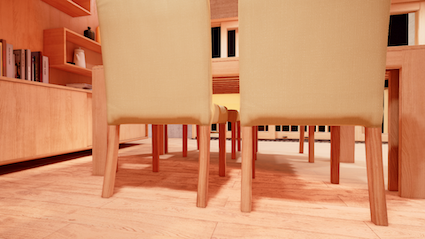}\label{fig:home_l}
	}
	\subfloat[]
	{
		\adjincludegraphics[width=0.175\linewidth,trim={{.2\width} 0 {.1\width} 0},clip]{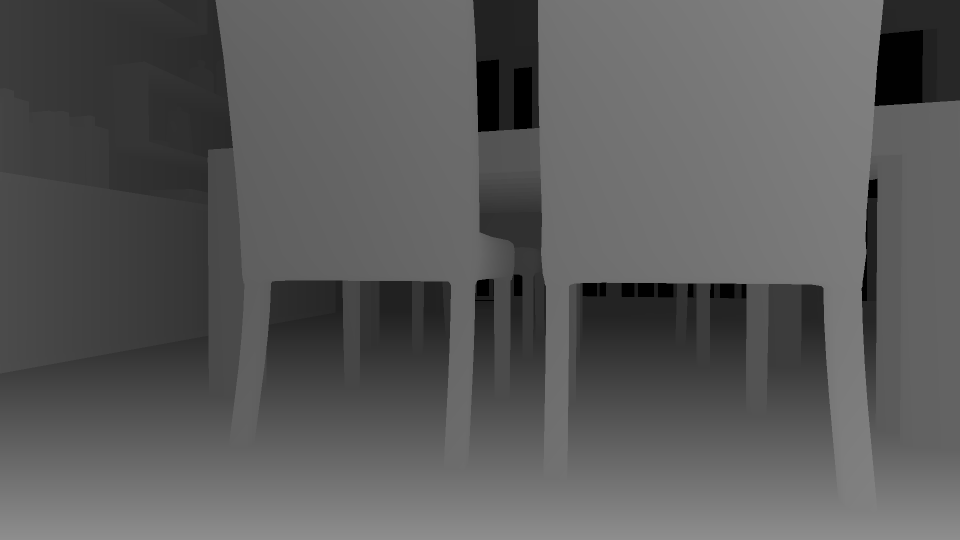}\label{fig:home_d_gt}
	}
	\subfloat[]
	{
		\adjincludegraphics[width=0.175\linewidth,trim={{.2\width} 0 {.1\width} 0},clip]{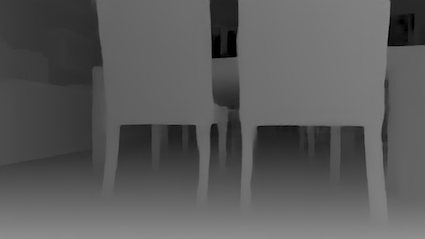}\label{fig:home_d}
	}
	\subfloat[]
	{
		\adjincludegraphics[width=0.175\linewidth,trim={{.2\width} 0 {.1\width} 0},clip]{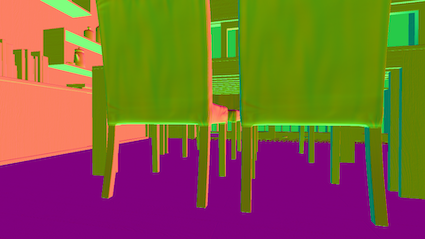}\label{fig:home_n_gt}
	}
	\subfloat[]
	{
		\adjincludegraphics[width=0.175\linewidth,trim={{.2\width} 0 {.1\width} 0},clip]{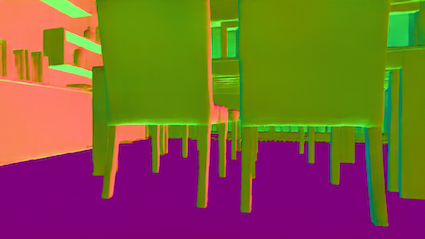}\label{fig:home_n}
	}
	\vspace{-0.8 em}
	\subfloat[]
	{
		\adjincludegraphics[width=0.175\linewidth,trim={{.2\width} 0 {.1\width} 0},clip]{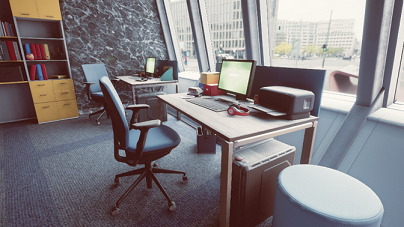}\label{fig:office_l}
	}
	\subfloat[]
	{
		\adjincludegraphics[width=0.175\linewidth,trim={{.2\width} 0 {.1\width} 0},clip]{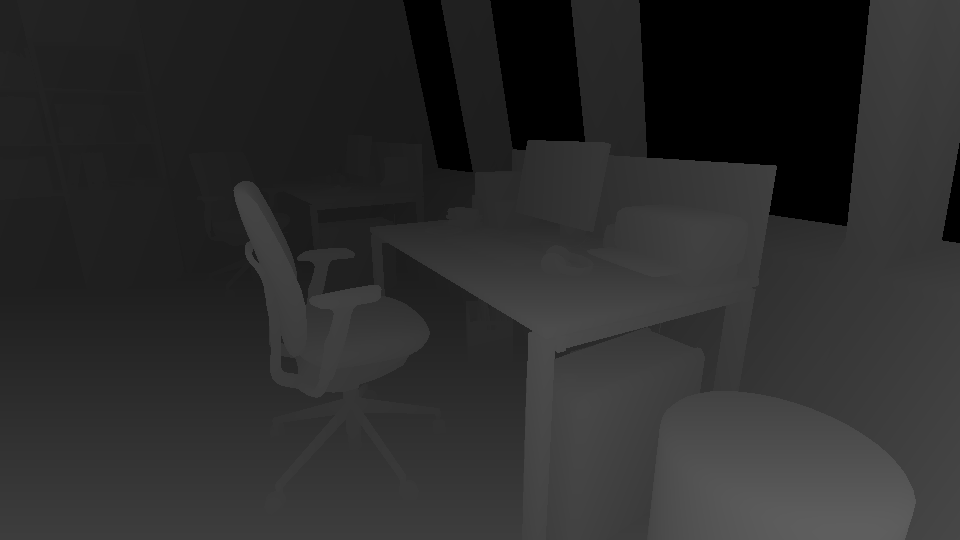}\label{fig:office_d_gt}
	}
	\subfloat[]
	{
		\adjincludegraphics[width=0.175\linewidth,trim={{.2\width} 0 {.1\width} 0},clip]{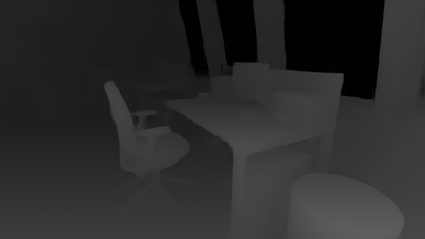}\label{fig:office_d}
	}
	\subfloat[]
	{
		\adjincludegraphics[width=0.175\linewidth,trim={{.2\width} 0 {.1\width} 0},clip]{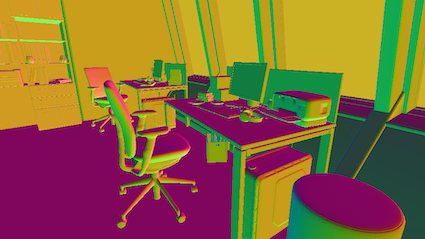}\label{fig:office_n_gt}
	}
	\subfloat[]
	{
		\adjincludegraphics[width=0.175\linewidth,trim={{.2\width} 0 {.1\width} 0},clip]{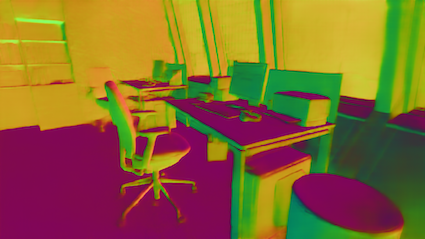}\label{fig:office_n}
	}
	\vspace{-0.8 em}
	\subfloat[]
	{
		\adjincludegraphics[width=0.175\linewidth,trim={{.2\width} 0 {.1\width} 0},clip]{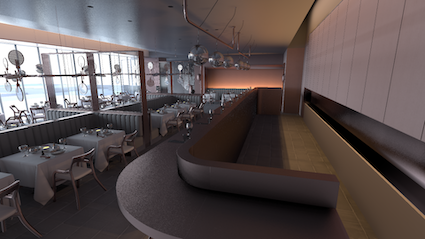}\label{fig:rest_l}
	}
	\subfloat[]
	{
		\adjincludegraphics[width=0.175\linewidth,trim={{.2\width} 0 {.1\width} 0},clip]{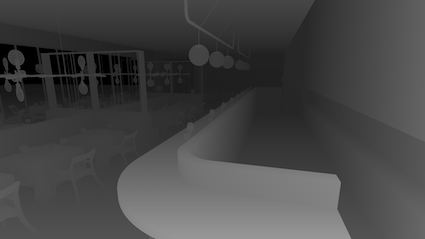}\label{fig:rest_d_gt}
	}
	\subfloat[]
	{
		\adjincludegraphics[width=0.175\linewidth,trim={{.2\width} 0 {.1\width} 0},clip]{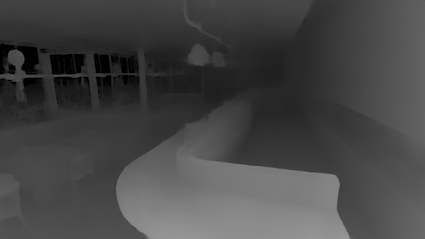}\label{fig:rest_d}
	}
	\subfloat[]
	{
		\adjincludegraphics[width=0.175\linewidth,trim={{.2\width} 0 {.1\width} 0},clip]{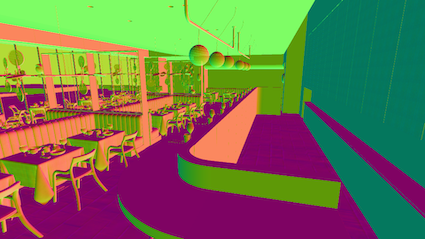}\label{fig:rest_n_gt}
	}
	\subfloat[]
	{
		\adjincludegraphics[width=0.175\linewidth,trim={{.2\width} 0 {.1\width} 0},clip]{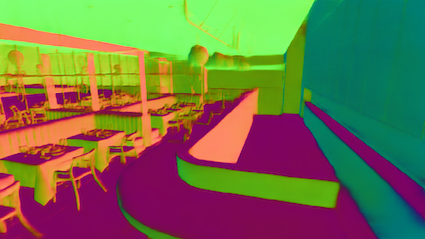}\label{fig:rest_n}
	}
	\vspace{-0.8 em}
	\subfloat[Left View]
	{
		\adjincludegraphics[width=0.175\linewidth,trim={{.2\width} 0 {.1\width} 0},clip]{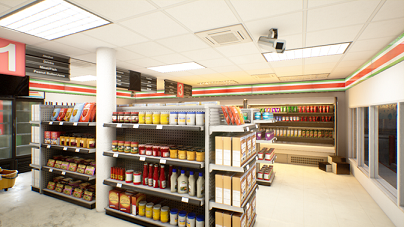}\label{fig:store_l}
	}
	\subfloat[Disparity GT]
	{
		\adjincludegraphics[width=0.175\linewidth,trim={{.2\width} 0 {.1\width} 0},clip]{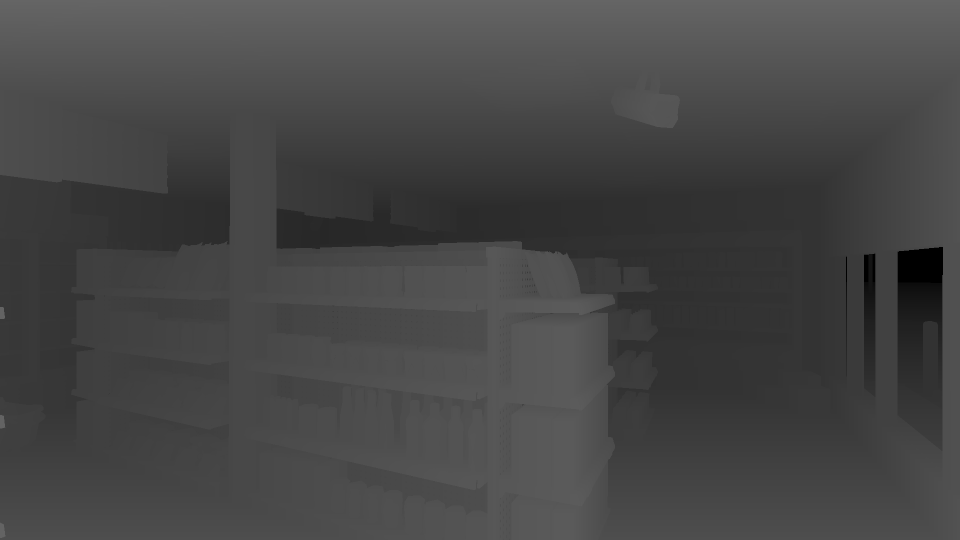}\label{fig:store_d_gt}
	}
	\subfloat[FADNet]
	{
		\adjincludegraphics[width=0.175\linewidth,trim={{.2\width} 0 {.1\width} 0},clip]{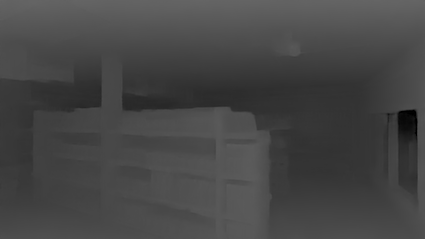}\label{fig:store_d}
	}
	\subfloat[Normal GT]
	{
		\adjincludegraphics[width=0.175\linewidth,trim={{.2\width} 0 {.1\width} 0},clip]{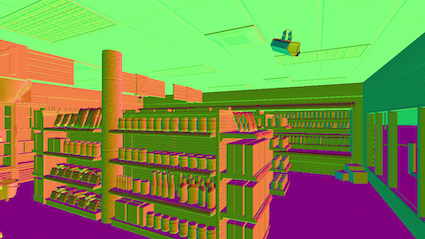}\label{fig:store_n_gt}
	}
	\subfloat[DTN-Net]
	{
		\adjincludegraphics[width=0.175\linewidth,trim={{.2\width} 0 {.1\width} 0},clip]{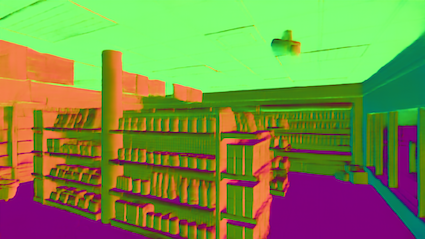}\label{fig:store_n}
	}
	\caption{Disparity Prediction Results on IRS Images.}
	\label{fig:dn_on_irs}
\end{figure}

Fig. \ref{fig:dn_on_irs} shows some predicted disparity and surface normal maps of IRS samples. It is observed that FADNet and DTN-Net learns very decent disparity and surface normal information close to the ground truth. Generally the models trained on our IRS can have better performance on the indoor images, compared to other models trained on FlyingThings3D. This is also visible in the provided video.

\begin{figure}[htbp]
	\captionsetup[subfigure]{labelformat=empty, farskip=0pt}
	\centering
	\subfloat[]
	{
		\adjincludegraphics[width=0.3\linewidth,trim={{.1\width} 0 {.1\width} 0},clip]{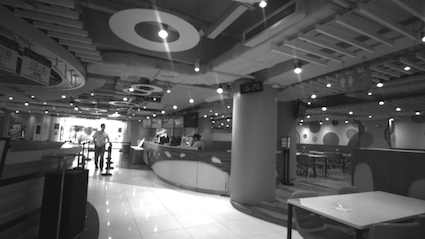}\label{fig:rc_l}
	}
	\subfloat[]
	{
		\adjincludegraphics[width=0.3\linewidth,trim={{.1\width} 0 {.1\width} 0},clip]{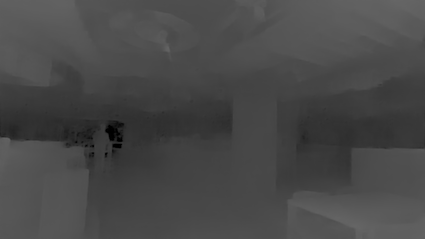}\label{fig:rc_d}
	}
	\subfloat[]
	{
		\adjincludegraphics[width=0.3\linewidth,trim={{.1\width} 0 {.1\width} 0},clip]{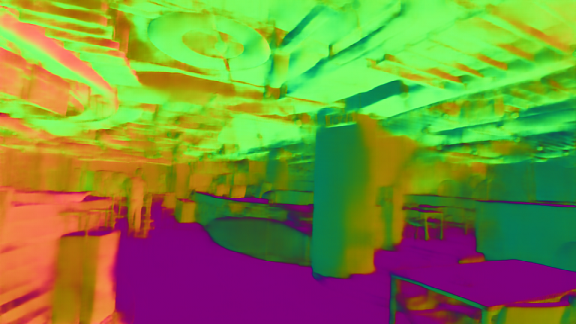}\label{fig:rc_n}
	}
	\vspace{-0.8 em}
	\subfloat[Left View]
	{
		\adjincludegraphics[width=0.3\linewidth,trim={{.1\width} 0 {.1\width} 0},clip]{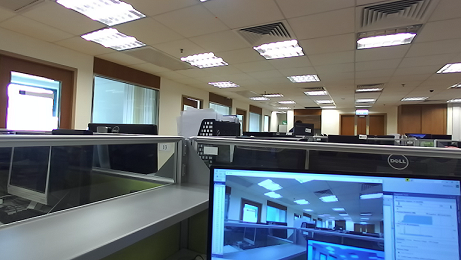}\label{fig:ro_l}
	}
	\subfloat[FADNet]
	{
		\adjincludegraphics[width=0.3\linewidth,trim={{.1\width} 0 {.1\width} 0},clip]{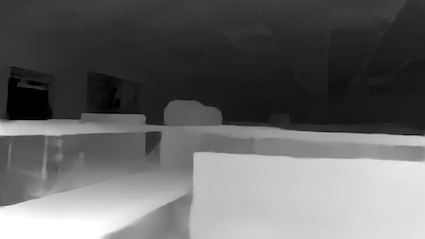}\label{fig:ro_d}
	}
	\subfloat[DTN-Net]
	{
		\adjincludegraphics[width=0.3\linewidth,trim={{.1\width} 0 {.1\width} 0},clip]{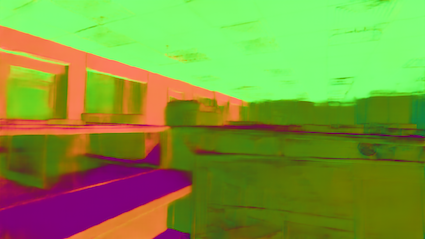}\label{fig:ro_n}
	}
	\caption{Disparity Prediction Results on Real World Images.}
	\label{fig:dn_on_real}
\end{figure}

Fig. \ref{fig:dn_on_real} shows two real-world captured examples. Two rows respectively report a canteen scene and an office scene. Although the prediction may fail on some far small objects and boundries, FADNet and DTN-Net still achieve considerable results in aspects of smooth planes and reasonable orientation of them. 
\subsection{Point Cloud Reconstruction}\label{sec:pcd}
\begin{figure*}[htbp]
	\captionsetup[subfigure]{labelformat=empty, farskip=0pt}
	\centering
	\subfloat[]
	{
		\adjincludegraphics[width=0.3\linewidth,trim={{.2\width} 0 {.1\width} 0},clip]{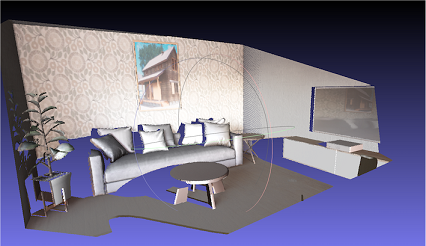}\label{fig:home_pcd_gt}
	}
	\subfloat[]
	{
		\adjincludegraphics[width=0.3\linewidth,trim={{.2\width} 0 {.1\width} 0},clip]{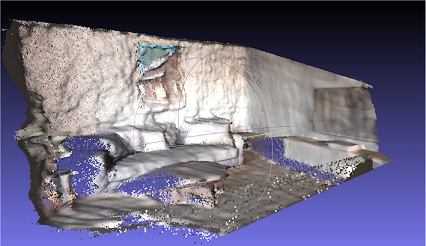}\label{fig:home_pcd_rdnet}
	}
	\subfloat[]
	{
		\adjincludegraphics[width=0.3\linewidth,trim={{.2\width} 0 {.1\width} 0},clip]{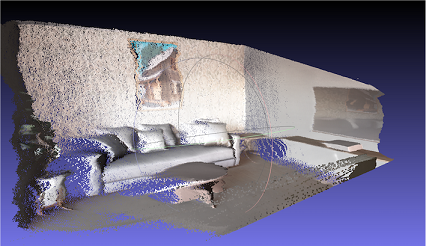}\label{fig:home_pcd_dtonnet}
	}
	\vspace{-0.8 em}
	\subfloat[GT]
	{
		\adjincludegraphics[width=0.3\linewidth,trim={{.2\width} 0 {.1\width} 0},clip]{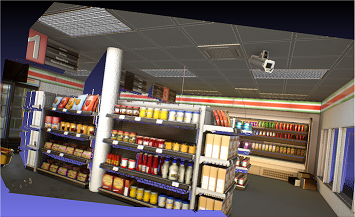}\label{fig:store_pcd_gt}
	}
	\subfloat[FADNet]
	{
		\adjincludegraphics[width=0.3\linewidth,trim={{.2\width} 0 {.1\width} 0},clip]{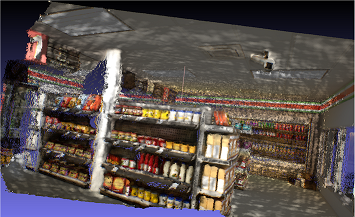}\label{fig:store_pcd_rdnet}
	}
	\subfloat[DTN-Net]
	{
		\adjincludegraphics[width=0.3\linewidth,trim={{.2\width} 0 {.1\width} 0},clip]{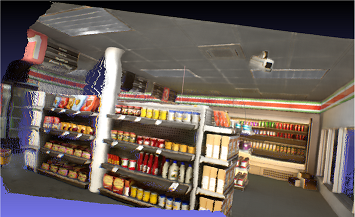}\label{fig:store_pcd_dtonnet}
	}
	\caption{Point Cloud Reconstruction by Predicted Disparity and Surface Normal on IRS Samples. The first column shows the point cloud generated from the ground truth of disparity and surface normal. The second column shows the point cloud generated from the predicted disparity by FADNet. The third column shows the point cloud generated from the predicted disparity and surface normal by DTN-Net. }
	\label{fig:pcd_on_irs}
\end{figure*}

\begin{figure*}[htbp]
	\captionsetup[subfigure]{labelformat=empty, farskip=0pt}
	\centering
	\subfloat[]
	{
		\adjincludegraphics[width=0.3\linewidth,trim={{.2\width} 0 {.1\width} 0},clip]{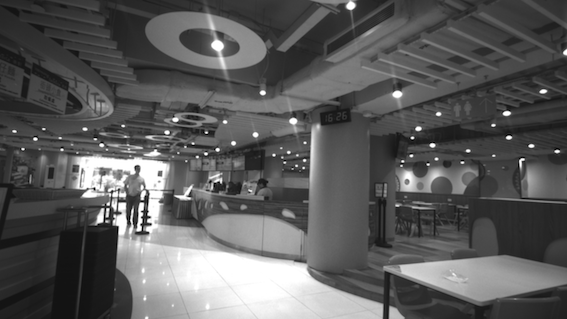}\label{fig:real_canteen_l}
	}
	\subfloat[]
	{
		\adjincludegraphics[width=0.3\linewidth,trim={{.2\width} {.075\height} {.1\width} 0},clip]{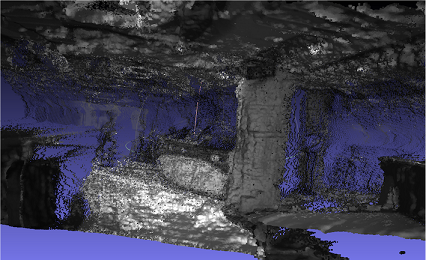}\label{fig:real_canteen_pcd_rdnet}
	}
	\subfloat[]
	{
		\adjincludegraphics[width=0.3\linewidth,trim={{.2\width} {.07\height} {.1\width} 0},clip]{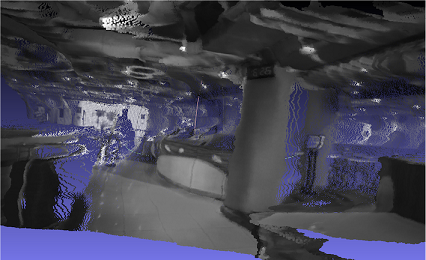}\label{fig:real_canteen_pcd_dtonnet}
	}
	\vspace{-0.8 em}
	\subfloat[Left View]
	{
		\adjincludegraphics[width=0.3\linewidth,trim={{.2\width} 0 {.1\width} 0},clip]{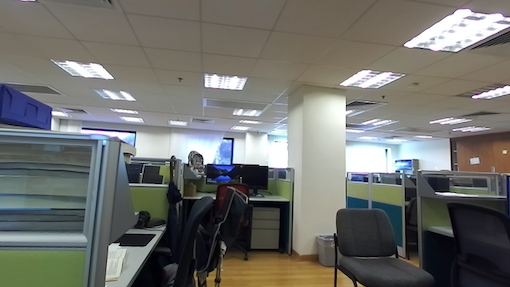}\label{fig:real_office_l}
	}
	\subfloat[FADNet]
	{
		\adjincludegraphics[width=0.3\linewidth,trim={{.2\width} {.075\height} {.1\width} 0},clip]{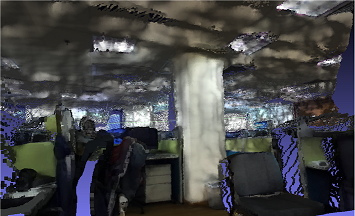}\label{fig:real_office_pcd_rdnet}
	}
	\subfloat[DTN-Net]
	{
		\adjincludegraphics[width=0.3\linewidth,trim={{.2\width} {.075\height} {.1\width} 0},clip]{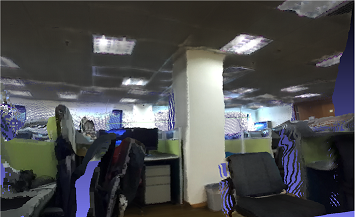}\label{fig:real_office_pcd_dtonnet}
	}
	\caption{Point Cloud Reconstruction by Predicted Disparity and Surface Normal on Real-World Captured Images. The first column shows the left view image. The second column shows the point cloud generated from the predicted disparity by FADNet. The third column shows the point cloud generated from the predicted disparity and surface normal by DTN-Net. }
	\label{fig:pcd_on_real}
\end{figure*}

To illustrate the usage of FADNet and DTN-Net for 3D reconstruction, we convert the predicted disparity maps into point clouds using Open3D library \footnote{\url{http://www.open3d.org}}, as shown in Fig. \ref{fig:pcd_on_irs} and Fig. \ref{fig:pcd_on_real}. For DTN-Net, we assign the predicted surface normal information to the generated point clouds. For FADNet, which does not predict the surface normal maps, Open3D API generates the surface information for them according to its estimated disparity.

Fig. \ref{fig:pcd_on_irs} shows the point cloud of two IRS samples, one room scene and one store scene. Overall speaking, the results of DTN-Net is much closer to groundtruth than those of FADNet. Since FADNet is trained with only the disparity information, it may miss the consistency between the disparity and the surface normal. However, DTN-Net learns the disparity and surface normal synchronously, which helps generate smooth planes, such as floor, ceiling and wall, and continuous object surface. This kind of smoothness results in precise light reflection behaviors. 

Fig. \ref{fig:pcd_on_real} shows the point cloud of two real-world captured samples, one canteen scene and one office scene. We observe that there are lots of black points in the results of FADNet, which are caused by the wrong light reflection. Thus, the wall, ceiling and floor in the point cloud of FADNet show uneven surface instead of flat planes, while DTN-Net produces much better results. 

%
%


\end{document}